\documentclass[10pt,conference]{IEEEtran}
\IEEEoverridecommandlockouts
\usepackage{adjustbox}
\usepackage{algorithm}
\usepackage{algpseudocode}
\usepackage{amsfonts}
\usepackage{amsmath}
\usepackage{amssymb}
\usepackage{array}
\usepackage{balance}
\usepackage{booktabs}
\usepackage{colortbl}
\usepackage{diagbox}
\usepackage{float}
\usepackage{graphicx}
\usepackage{hyperref}
\usepackage[utf8]{inputenc}
\usepackage{lipsum}
\usepackage{makecell}
\usepackage{multirow}
\usepackage{pifont}
\usepackage{pgf-pie}
\usepackage{pgfplots}
\usepackage{pgfplotstable}
\usepackage{slashbox}
\usepackage{tabularx}
\usepackage{textcomp}
\usepackage{tcolorbox}
\usepackage{tikz}
\usepackage[table,dvipsnames]{xcolor}
\usepackage{xurl}
\pgfplotsset{compat=1.18}
\usepgfplotslibrary{polar}
\usetikzlibrary{patterns,shapes.geometric,arrows,arrows.meta,positioning,calc,fit,backgrounds}
\newcommand{\memtype}[1]{\texttt{\small #1}}
\definecolor{navyblue}{RGB}{31,56,100}
\definecolor{medblue}{RGB}{46,95,163}
\definecolor{lightblue}{RGB}{214,228,240}
\definecolor{colorVec}{RGB}{15,110,86}
\definecolor{colorHyb}{RGB}{180,100,10}
\definecolor{colorOth}{RGB}{100,100,110}
\definecolor{memred}{RGB}{163,45,45}
\definecolor{memgreen}{RGB}{15,110,86}
\definecolor{memamber}{RGB}{180,117,23}
\definecolor{rowgray}{RGB}{245,245,245}

\newcommand{\sys}[1]{\textsc{#1}}

\newcommand{\ck}{\textcolor{memgreen}{\ding{51}}}
\newcommand{\cx}{\textcolor{memred}{\ding{55}}}
\newcommand{\cm}{\textcolor{memamber}{$\sim$}}
\newcolumntype{L}[1]{>{\raggedright\arraybackslash}p{#1}}
\newcolumntype{C}[1]{>{\centering\arraybackslash}p{#1}}
\newcolumntype{R}[1]{>{\raggedleft\arraybackslash}p{#1}}

\begin{document}
\title{Memanto: Typed Semantic Memory with Information-Theoretic Retrieval for Long-Horizon Agents}

%% ── Authors ────────────────────────────────────────────
\makeatletter
\newcommand{\linebreakand}{%
  \end{@IEEEauthorhalign}
  \hfill\mbox{}\par
  \mbox{}\hfill\begin{@IEEEauthorhalign}
}
\author{
    \IEEEauthorblockN{Seyed Moein Abtahi}
    \IEEEauthorblockA{%
    \textit{Moorcheh AI}\\
    \textit{EdgeAI Innovations}\\
    seyedmoein.abtahi@ontariotechu.net}
    \and
    \IEEEauthorblockN{Rasa Rahnema}
    \IEEEauthorblockA{
    \textit{Moorcheh AI}\\
    \textit{EdgeAI Innovations}\\
    ryrahnem@uwaterloo.ca}
    \and
    \IEEEauthorblockN{Hetkumar Patel}
    \IEEEauthorblockA{%
    \textit{Moorcheh AI}\\
    \textit{EdgeAI Innovations}\\
    hetkumardineshbhai.patel@sheridancollege.ca}
    % --- USE THE CUSTOM COMMAND HERE ---
    \linebreakand 
    \IEEEauthorblockN{Neel Patel}
    \IEEEauthorblockA{%
    \textit{Moorcheh AI}\\
    \textit{EdgeAI Innovations}\\
    neel@edgeaiinnovations.com}
    \and
    \IEEEauthorblockN{Majid Fekri}
    \IEEEauthorblockA{%
    \textit{Moorcheh AI}\\
    \textit{EdgeAI Innovations}\\
    majid.fekri@edgeaiinnovations.com}
    \and
    \IEEEauthorblockN{Tara Khani}
    \IEEEauthorblockA{%
    \textit{Moorcheh AI}\\
    \textit{EdgeAI Innovations}\\
    tara.khani@edgeaiinnovations.com}
}
%% ── Chapters ────────────────────────────────────────────
\maketitle
\begin{abstract}
The transition from stateless language model inference to persistent, multi session autonomous agents has revealed memory to be a primary architectural bottleneck in the deployment of production grade agentic systems. Existing methodologies largely depend on hybrid semantic graph architectures, which impose substantial computational overhead during both ingestion and retrieval. These systems typically require large language model mediated entity extraction, explicit graph schema maintenance, and multi query retrieval pipelines. This paper introduces Memanto, a universal memory layer for agentic artificial intelligence that challenges the prevailing assumption that knowledge graph complexity is necessary to achieve high fidelity agent memory. Memanto integrates a typed semantic memory schema comprising thirteen predefined memory categories, an automated conflict resolution mechanism, and temporal versioning. These components are enabled by Moorcheh's Information Theoretic Search engine, a no indexing semantic database that provides deterministic retrieval within sub ninety millisecond latency while eliminating ingestion delay. Through systematic benchmarking on the LongMemEval and LoCoMo evaluation suites, Memanto achieves state of the art accuracy scores of 89.8 percent and 87.1 percent respectively. These results surpass all evaluated hybrid graph and vector based systems while requiring only a single retrieval query, incurring no ingestion cost, and maintaining substantially lower operational complexity. A five stage progressive ablation study is presented to quantify the contribution of each architectural component, followed by a discussion of the implications for scalable deployment of agentic memory systems.
\end{abstract}

\section{Introduction}
\label{sec:intro}

Memory has become a fundamental architectural component in the design of foundation model based agents \cite{hu2026memoryageaiagents,v2026agenticartificialintelligenceai}. As large language models transition from single turn question answering systems to autonomous agents capable of multi step reasoning, tool utilization, and long horizon task execution \cite{yao2023react,wang2024survey}, their inherent limitation, namely the lack of persistent state across sessions, has emerged as a central engineering challenge in the development of agentic artificial intelligence systems.

Industry projections indicate that the agentic artificial intelligence market will expand from 7.8 billion dollars to more than 52 billion dollars by 2030. Gartner further estimates that 40 percent of enterprise applications will incorporate AI agents by the end of 2026, compared to less than 5 percent in 2025. This accelerated adoption has generated a critical need for memory infrastructure that satisfies production requirements, including high accuracy, low latency, cost efficiency, and reduced operational complexity.

Current approaches within the field remain heterogeneous and increasingly complex. A range of frameworks, including Mem0 \cite{chhikara2025mem0buildingproductionreadyai}, Zep \cite{rasmussen2025zep}, Letta \cite{packer2024memgptllmsoperatingsystems}, and A-MEM 
\cite{xu2025amem}, propose architectures that integrate knowledge graphs, temporal graph databases, multi query retrieval strategies, and recursive large language model driven ingestion pipelines. Although these systems demonstrate competitive benchmark performance, they introduce substantial computational and operational overhead. This paper characterizes this phenomenon as the \textbf{``Memory Tax''}, defined as the cumulative increase in compute cost, latency, and system complexity associated with memory ingestion and retrieval processes.

Memanto, built upon Moorcheh's Information Theoretic Search engine, a no indexing semantic database based on Information Theoretic Vector Compression, demonstrates that highly optimized semantic retrieval combined with structured memory typing and automated conflict resolution can achieve and surpass the performance of hybrid graph and vector architectures. This is accomplished while eliminating ingestion overhead, reducing retrieval to a single query, and removing the need for schema management. Fig. \ref{fig:timeline} contextualizes Memanto within the broader evolution of memory systems for agentic artificial intelligence.

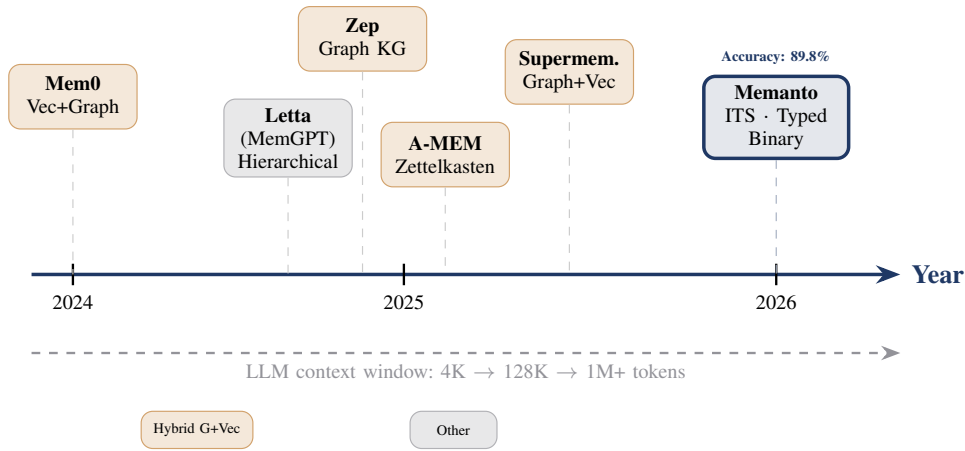
\begin{figure*}[ht!]
\centering
\scalebox{1.1}{
\begin{tikzpicture}[font=\small,
  sysbox/.style={draw,rounded corners=4pt,align=center,minimum width=1.55cm,minimum height=0.78cm,inner sep=3pt,font=\scriptsize},
  vecbox/.style={sysbox,fill=colorVec!15,draw=colorVec!60},
  hybbox/.style={sysbox,fill=colorHyb!15,draw=colorHyb!60},
  othbox/.style={sysbox,fill=colorOth!15,draw=colorOth!50},
  membox/.style={sysbox,very thick,fill=navyblue!12,draw=navyblue,minimum width=1.75cm,minimum height=0.92cm},
]
\draw[-{Stealth},very thick,navyblue] (4.5,0)--(15.,0) node[right,font=\small\bfseries]{Year};
\foreach \x/\yr in {5/2024,9/2025,13.5/2026}{
  \draw[thick](\x,0.13)--(\x,-0.13);
  \node[font=\scriptsize,below]at(\x,-0.13){\yr};
}
\node[othbox](letta)at(7.6,1.65){\textbf{Letta}\\(MemGPT)\\Hierarchical};
\draw[gray!40,thin,dashed](letta.south)--(7.6,0);
\node[hybbox](mem0)at(5.0,2.15){\textbf{Mem0}\\Vec+Graph};
\draw[gray!40,thin,dashed](mem0.south)--(5.0,0);
\node[hybbox](zep)at(8.5,2.85){\textbf{Zep}\\Graph KG};
\draw[gray!40,thin,dashed](zep.south)--(8.5,0);
\node[hybbox](amem)at(9.5,1.45){\textbf{A-MEM}\\Zettelkasten};
\draw[gray!40,thin,dashed](amem.south)--(9.5,0);
\node[hybbox](super)at(11.0,2.45){\textbf{Supermem.}\\Graph+Vec};
\draw[gray!40,thin,dashed](super.south)--(11.0,0);
\node[membox,label={[navyblue,font=\tiny\bfseries]above:Accuracy: 89.8\%}](memanto)at(13.5,1.9){\textbf{Memanto}\\ITS $\cdot$ Typed\\Binary};
\draw[navyblue!40,thin,dashed](memanto.south)--(13.5,0);
\draw[-{Stealth},dashed,colorOth!70,thick] (4.5,-0.95)--(15.0,-0.95) node[midway,below,font=\scriptsize]{LLM context window: 4K $\rightarrow$ 128K $\rightarrow$ 1M+ tokens};
% \node[vecbox,minimum width=1.1cm,minimum height=0.45cm,font=\tiny] at(3.0,-1.88){Vector only};
\node[hybbox,minimum width=1.35cm,minimum height=0.45cm,font=\tiny] at(6.5,-1.88){Hybrid G+Vec};
\node[othbox,minimum width=1.05cm,minimum height=0.45cm,font=\tiny] at(9.6,-1.88){Other};
\end{tikzpicture}
}
\caption{Evolution of agentic memory systems (2023 to 2026). Despite rapidly growing LLM context windows, architectures have grown more complex. Memanto (2026) achieves SOTA accuracy using a simpler, vector only approach that eliminates graph infrastructure and LLM mediated ingestion entirely.}
\label{fig:timeline}
\end{figure*}

\subsection{Contributions}

\begin{enumerate}
\item \textbf{Architectural:} We introduce a production grade agentic memory system that achieves state of the art benchmark performance using a vector based architecture with zero cost ingestion. The system incorporates a typed memory schema consisting of thirteen semantic categories and an integrated conflict resolution mechanism, without reliance on knowledge graphs, multi query retrieval, or large language model mediated ingestion.

\item \textbf{Empirical:} We present a five stage progressive ablation study conducted on LongMemEval \cite{wu2024longmemeval} and LoCoMo \cite{maharana2024locomo}. The study evaluates the effects of retrieval limit tuning, similarity threshold calibration, prompt design, and inference model selection. Final results of 89.8 percent and 87.1 percent respectively establish a new state of the art among vector based systems.

\item \textbf{Systems:} We formally analyze and quantify the Memory Tax associated with hybrid graph and vector architectures. Our results demonstrate that the additional overhead yields diminishing performance gains when compared to optimized semantic retrieval, particularly under deterministic exact match search rather than approximate nearest neighbor methods.

\item \textbf{Design Principles:} We propose six design principles for production ready agentic memory systems, derived from systematic evaluation of agent requirements and informed by feedback from real world deployments.
\end{enumerate}
\section{Background and Related Work}\label{sec:related}

This section situates the proposed approach within the broader intellectual and systems landscape of agentic memory. We begin with cognitive foundations that inform modern memory abstractions, followed by recent taxonomies that attempt to organize the rapidly evolving design space. We then examine dominant architectural paradigms, with particular emphasis on hybrid graph based systems, before analyzing the indexing and ingestion bottlenecks that constrain current implementations. Finally, we review the principal evaluation benchmarks used to assess long term memory in agentic systems.

\subsection{Cognitive Foundations of Memory Taxonomy}

Cognitive science provides a principled framework for structuring memory in artificial agents. Tulving's foundational work \cite{tulving1972episodic} distinguishes between episodic memory, which captures event specific and temporally situated experiences, semantic memory, which encodes general knowledge and factual information, and procedural memory, which governs skills and behavioral rules. These distinctions have become directly relevant to the design of memory systems for large language model based agents.

Baddeley's working memory model \cite{baddeley1992working}, consisting of the phonological loop, visuospatial sketchpad, and central executive, exhibits a strong conceptual correspondence with modern retrieval augmented generation architectures. In this mapping, the phonological loop aligns with in context token buffers, the visuospatial sketchpad with structured retrieval representations, and the central executive with the reasoning and control mechanisms of the agent.

Recent work has emphasized the importance of episodic memory for long horizon agent behavior. MacPherson et al. \cite{macpherson2025episodic} argue that episodic representations enable temporal specificity and contextual binding that cannot be achieved through semantic retrieval alone. ENGRAM \cite{patel2025engram} operationalizes this insight by implementing three distinct memory types with a unified routing and retrieval mechanism, demonstrating that typed memory separation significantly improves performance on both \sys{LoCoMo} and \sys{LongMemEval}. Memanto extends this principle through a more granular schema comprising thirteen memory categories.

\subsection{Memory Surveys and Taxonomies (2024 to 2026)}

A growing body of survey literature seeks to impose structure on the increasingly heterogeneous memory landscape. Zhang et al. \cite{hu2026memoryageaiagents} categorize memory systems along three dimensions: forms, functions, and dynamics, identifying token level, parametric, and latent memory as primary representations. Abou Ali et al. \cite{Abou_Ali_2025} propose a dual paradigm framework that distinguishes symbolic or classical approaches from neural or generative paradigms. 

Arunkumar et al. \cite{v2026agenticartificialintelligenceai} describe a four layer architecture encompassing perception, memory, the agent core, and action. Nisa et al. \cite{NISA202669} position memory as the substrate enabling coherent reasoning and planning across time, while Wang et al. \cite{wang2024survey} identify long term memory as a central unresolved challenge in large language model based agents. Sumers et al. \cite{sumers2023cognitive} further formalize the correspondence between cognitive architectures and agent system design, mapping perception, memory, learning, and decision making to computational components.

\subsection{Knowledge Graph Based Memory Systems (2024 to 2026)}

The dominant paradigm for production grade agent memory systems has converged on hybrid architectures that integrate dense vector representations with structured knowledge graphs.

\textbf{MemGPT / Letta} \cite{packer2024memgptllmsoperatingsystems} introduces a virtual memory abstraction inspired by operating systems, in which information is dynamically paged between context and external storage. While conceptually influential, this approach relies on recursive summarization and hierarchical compression, which can introduce latency variability and loss of information fidelity, particularly when precise textual recall is required.

\textbf{Mem0} \cite{chhikara2025mem0buildingproductionreadyai} implements a three tier memory hierarchy spanning user, session, and agent scopes. The system combines vector retrieval, graph based relational storage, and key value indexing. Although it demonstrates strong empirical performance, its ablation results indicate that the graph augmented variant yields only marginal improvements over the base vector configuration. This raises questions regarding the necessity of graph infrastructure relative to its associated computational and operational costs.

In practice, each memory insertion in the graph augmented configuration triggers a synchronous multi stage pipeline consisting of large language model driven entity extraction, vector embedding and index updates, and graph synchronization. This process transforms low latency write operations into multi second procedures, thereby exemplifying the accumulation of computational overhead associated with complex memory architectures.

\textbf{Zep / Graphiti} \cite{rasmussen2025zep} extends the graph based paradigm by incorporating temporal versioning and bi temporal indexing to support enterprise grade auditing and compliance requirements. However, the reliance on synchronous extraction pipelines introduces ingestion latency, delaying the availability of newly stored information for retrieval.

\textbf{A-MEM} \cite{xu2025amem} adopts a Zettelkasten inspired design in which memories are represented as interconnected notes enriched with contextual metadata. While this enables associative retrieval, it requires a full inference step for each memory insertion, increasing both latency and cost.

\textbf{Hindsight} \cite{hindsight2025} and subsequent reflective memory frameworks \cite{le2025reflective} achieve high benchmark accuracy through multi stage retrieval and reflection mechanisms. These systems rely on parallel queries and iterative reasoning passes, resulting in significantly higher system complexity relative to single query retrieval approaches.

Emerging evidence challenges the necessity of such architectural complexity. Merrill et al. \cite{merrill2026evaluating} demonstrate that comparatively simple retrieval based systems can outperform more elaborate memory hierarchies on existing benchmarks, suggesting that current evaluation protocols may not fully capture the benefits of structured memory organization.

\subsection{The Indexing and Ingestion Bottleneck}

Retrieval augmented generation \cite{lewis2020rag} establishes the canonical paradigm for augmenting language models with external memory. Subsequent systems extend this framework through hierarchical memory abstractions and hybrid storage mechanisms. However, traditional vector databases rely on approximate nearest neighbor indexing structures such as hierarchical navigable small world graphs \cite{malkov2018hnsw}, which introduce non negligible delays between data ingestion and query availability.

For agentic systems operating in interactive or iterative settings, this delay is problematic. An agent may need to store information and immediately retrieve it within the same reasoning trajectory. Any latency in indexing directly impairs this capability.

\sys{LongMemEval} \cite{wu2024longmemeval} provides a structured analysis of memory system design, decomposing performance into indexing, retrieval, and reading stages. The study identifies key factors including granularity of stored information, key construction, query formulation, and reading strategies. Empirical results demonstrate that fine grained session decomposition, enriched key representations, temporally aware query expansion, and structured reading techniques substantially improve accuracy.

Additional work highlights limitations in long context processing. Liu et al. \cite{liu2024lost} identify a degradation effect in which models exhibit reduced accuracy for information located in the middle portions of extended contexts. This finding reinforces the importance of targeted retrieval mechanisms that prioritize relevance over raw context length.

Alternative architectures such as HippoRAG \cite{hipporag2024} and RAPTOR \cite{raptor2024} address long range dependencies through hierarchical or graph based representations, but introduce additional system complexity. REPLUG \cite{shi2024replug} demonstrates that combining high recall retrieval with post retrieval verification improves robustness, aligning with the principle of prioritizing recall in memory systems.

\subsection{Evaluation Benchmarks}

\textbf{\sys{LongMemEval}} \cite{wu2024longmemeval} is a large scale benchmark comprising 500 curated questions spanning six categories, including user specific information, assistant responses, preferences, knowledge updates, temporal reasoning, and multi session interactions. The dataset is embedded within extended dialogues that can scale to over one million tokens across hundreds of sessions, providing a comprehensive test of long term memory capabilities.

\textbf{\sys{LoCoMo}} \cite{maharana2024locomo} consists of long form multi session dialogues with diverse reasoning requirements, including single hop, multi hop, open domain, and temporal queries. It serves as a complementary benchmark emphasizing conversational continuity and reasoning depth.

Additional benchmarks, including MemoryBank \cite{zhong2024memorybank}, PerLTQA \cite{du2024perltqa}, DialSim \cite{kim2024dialsim}, MemoryAgentBench \cite{memoryagentbench2025}, and long context evaluation frameworks \cite{terranova2025evaluatinglongtermmemorylongcontext}, further expand the evaluation landscape. However, recent analyses suggest that as model context windows increase, benchmark performance increasingly reflects underlying language model reasoning capabilities rather than the quality of the memory architecture itself, motivating the development of more targeted evaluation protocols.
\section{Design Principles and Architecture}
\label{sec:design}

Before describing Memanto's architecture, we articulate the six design principles that guided its development. These principles emerged from systematic analysis of agent operational requirements, including structured feedback from production AI agent deployments, and directly from the failure modes documented in the benchmark literature.

A contributing source was a structured dialogue with Claude (Anthropic)~\cite{claude2025}, in which the model was asked to articulate the limitations of its own memory architecture. The model identified passive context injection as the root failure mode, and independently surfaced seven specific gaps: the inability to query memory by relevance, the absence of temporal decay signals, the lack of confidence and provenance tagging, the flattening of episodic, semantic, and procedural memory into a single undifferentiated store, the absence of contradiction handling and versioning, the lack of scope and permissioning, and the absence of human-readable audit logs. This self-diagnosis maps directly onto desiderata D1 through D6, and informed both the framing and prioritization of Memanto's architectural requirements. The use of a frontier language model as a requirements elicitation source reflects a broader methodological principle: the systems most qualified to identify the failure modes of agent memory are the agents themselves. For more information, please visit Moorcheh.ai\footnote{\url{https://moorcheh.ai}} 
and Memanto.ai\footnote{\url{https://memanto.ai}}.

\subsection{Six Desiderata for Production Agentic Memory}

\noindent\textbf{D1. Queryable, not injectable.}
Agents need the ability to query memory based on relevance to the current task, not receive a static blob of context injected at conversation start. The distinction is between providing an agent with a pre-assembled dossier versus giving it a librarian it can consult on demand. Static injection fails when the injected context exceeds the context window, contains irrelevant content, or misses recently stored facts that are not yet in the injected snapshot.

\noindent\textbf{D2. Temporally aware with decay.}
Not all memories should have equal weight. A deadline mentioned yesterday carries different urgency than a preference stated six months ago. Memory must support temporal queries, versioning, and relevance decay signals that agents can reason about. This requirement maps directly to the Knowledge Update and Temporal Reasoning categories in LongMemEval \cite{wu2024longmemeval}, where systems without temporal awareness perform significantly below average.

\noindent\textbf{D3. Confidence and provenance tracking.}
A production memory system must distinguish between explicitly stated facts, inferred patterns, and potentially outdated information. Memory entries should carry provenance metadata that agents use to calibrate their confidence in retrieved context and avoid overconfident assertions on stale data.

\noindent\textbf{D4. Typed and hierarchical.}
Episodic memory (e.g., in our November conversation the user discussed X), semantic memory (e.g., the user is building a vector database startup), and procedural memory (e.g., when asked for reports the user prefers this format) serve fundamentally different retrieval purposes \cite{tulving1972episodic} and should be stored and queried with appropriate type semantics. 

\noindent\textbf{D5. Contradiction aware.}
When new information contradicts existing memory, the system must flag the conflict rather than silently overwrite or create inconsistency. For long-running agents, unresolved contradictions accumulate into what we term \emph{constraint drift}, a gradual erosion of the coherence of the agent's world model. MemoryAgentBench \cite{memoryagentbench2025} confirms that conflict resolution represents one of the most significant unsolved challenges, with all evaluated methods failing on multi-hop conflict scenarios.

\noindent\textbf{D6. Zero overhead ingestion.}
Every millisecond of ingestion latency is a millisecond where the agent cannot access its own recent experience. For real-time agentic workflows, memory must be available for retrieval at write time, with no indexing delay, no mandatory LLM extraction step, and no graph construction bottleneck.

Fig. \ref{fig:radar} visualises coverage across systems, while Table \ref{tab:desiderata} summarises it quantitatively.

\begin{figure}[!h]
\centering
\scalebox{1.}{
\begin{tikzpicture}[font=\scriptsize]
\begin{polaraxis}[
  width=7.2cm,
  xtick={0,60,120,180,240,300},
  xticklabels={D5 Conflict, D6 Ingest, D1 Query,
               D2 Temporal, D3 Provenance, D4 Typed},
  xticklabel style={font=\tiny},
  ymin=0, ymax=3,
  ytick={1,2,3},
  yticklabels={Low,Mid,High},
  yticklabel style={font=\tiny, anchor=north, rotate=0, yshift=-5pt, xshift = -4pt},
  grid=both, grid style={gray!25},
  x dir=reverse,
  legend style={at={(1.32,0.50)},anchor=west,font=\tiny,draw=gray!30,legend columns=1 , xshift= -1.6cm, yshift= -1.6cm},
]

%% order: D5, D6, D1, D2, D3, D4

\addplot[thick,navyblue,fill=navyblue,fill opacity=0.00,
         mark=*,mark size=1.5pt]
  coordinates{(0,3)(60,3)(120,3)(180,3)(240,3)(300,3)(360,3)};
\addlegendentry{Memanto}

\addplot[thick,colorHyb,fill=colorHyb,fill opacity=0.05,
         mark=square*,mark size=1.2pt,dashed]
  coordinates{(0,0.5)(60,0.5)(120,3)(180,2)(240,1)(300,1)(360,0.5)};
\addlegendentry{Mem0}

\addplot[thick,memred,fill=memred,fill opacity=0.05,
         mark=triangle*,mark size=1.2pt,dashed]
  coordinates{(0,0.5)(60,0.5)(120,3)(180,3)(240,1)(300,1)(360,0.5)};
\addlegendentry{Zep}

\addplot[thick,colorOth,fill=colorOth,fill opacity=0.10,
         mark=diamond*,mark size=1.2pt,dashed]
  coordinates{(0,0.5)(60,1)(120,3)(180,0.5)(240,0.5)(300,0.5)(360,0.5)};
\addlegendentry{Letta}

% \addplot[thick,memgreen,fill=memgreen,fill opacity=0.05,
%          mark=pentagon*,mark size=1.2pt,dashed]
%   coordinates{(0,0.5)(60,1)(120,3)(180,1)(240,0.5)(300,3)(360,0.5)};
% \addlegendentry{ENGRAM}

% \addplot[thick,orange,fill=orange,fill opacity=0.05,
%          mark=star,mark size=1.2pt,dashed]
%   coordinates{(0,0.5)(60,0.5)(120,3)(180,1)(240,1)(300,1)(360,0.5)};
% \addlegendentry{A-MEM}

\end{polaraxis}
\end{tikzpicture}
}
\caption{Desiderata coverage radar (D1 to D6). Memanto achieves full coverage across all six production desiderata. Key gaps: Zep and Mem0 are weak on D5--D6 (no conflict detection, high ingestion cost); Letta is weak on D2--D5; ENGRAM lacks D3, D5, and D6; A-MEM lacks D5--D6 with only partial D2--D4.}
\label{fig:radar}
\end{figure}
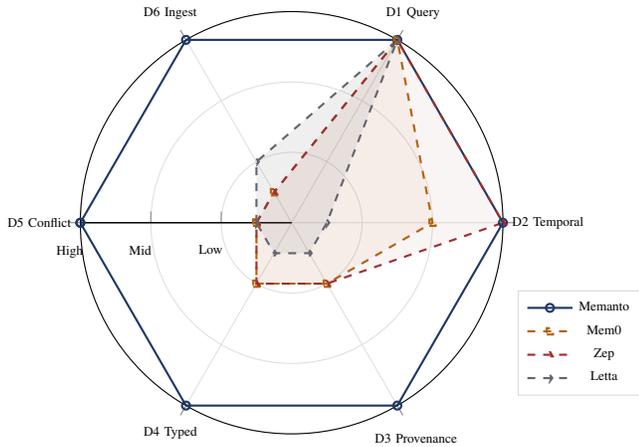

\begin{table}[!h]
\caption{Desiderata Coverage Across Agentic Memory Systems}
\label{tab:desiderata}
\centering\small
\scalebox{.9}{
\begin{tabular}{@{}L{1.92cm}C{0.55cm}C{0.62cm}C{0.72cm}C{0.55cm}C{0.62cm}C{0.62cm}@{}}
\toprule
\textbf{System}
  & \textbf{D1} & \textbf{D2} & \textbf{D3}
  & \textbf{D4} & \textbf{D5} & \textbf{D6} \\
\midrule
\rowcolor{lightblue}
\textbf{Memanto}
  &\ck&\ck&\ck&\ck&\ck&\ck\\
Mem0 \cite{chhikara2025mem0buildingproductionreadyai}
  &\ck&\ck&\cm&\cm&\cx&\cx\\
\rowcolor{rowgray}
Zep \cite{rasmussen2025zep}
  &\ck&\ck&\cm&\cm&\cx&\cx\\
% A-MEM \cite{xu2025amem}
%   &\ck&\cm&\cm&\cm&\cx&\cx\\
\rowcolor{rowgray}
% ENGRAM \cite{patel2025engram}
%   &\ck&\cm&\cx&\ck&\cx&\cm\\
Letta \cite{packer2024memgptllmsoperatingsystems}
  &\ck&\cx&\cx&\cx&\cx&\cm\\
\bottomrule
\multicolumn{7}{l}{\scriptsize \ck~Full support \quad \cm~Partial \quad \cx~Not supported}
\end{tabular}
}
\end{table}

\subsection{System Overview}

Memanto is designed as a local agentic platform that operates as a persistent FastAPI service, functioning as a dedicated memory agent in support of other AI agents. The system exposes three primary endpoints:

\begin{itemize}
    \item \textbf{/remember}: Commits items to memory with automatic typing, tagging, timestamping, conflict detection, and optional namespace scoping.
    \item \textbf{/recall}: Retrieves items from memory through Moorcheh's ITS-powered semantic search with configurable similarity thresholds and retrieval limits.
    \item \textbf{/answer}: Performs full Retrieval-Augmented Generation (RAG) with LLM intelligence applied on top of retrieved memory context.
\end{itemize}

Fig. \ref{fig:memanto-frontend} presents the Memanto Frontend Architecture, which comprises two layers. \textbf{(1) Agent Ecosystem.} IDE integrations, agent CLIs, custom agents built with Python, JavaScript, or LangChain, and a local web dashboard communicate with the gateway via CLI commands, REST API requests, and status controls. \textbf{(2) Memanto Gateway.} The gateway receives all incoming requests and routes them through two internal components: the Memanto CLI Engine, which handles command-line interactions, and the Memanto FastAPI Server, which serves REST API and dashboard traffic. All outbound calls from the gateway are forwarded to the backend shared services layer.

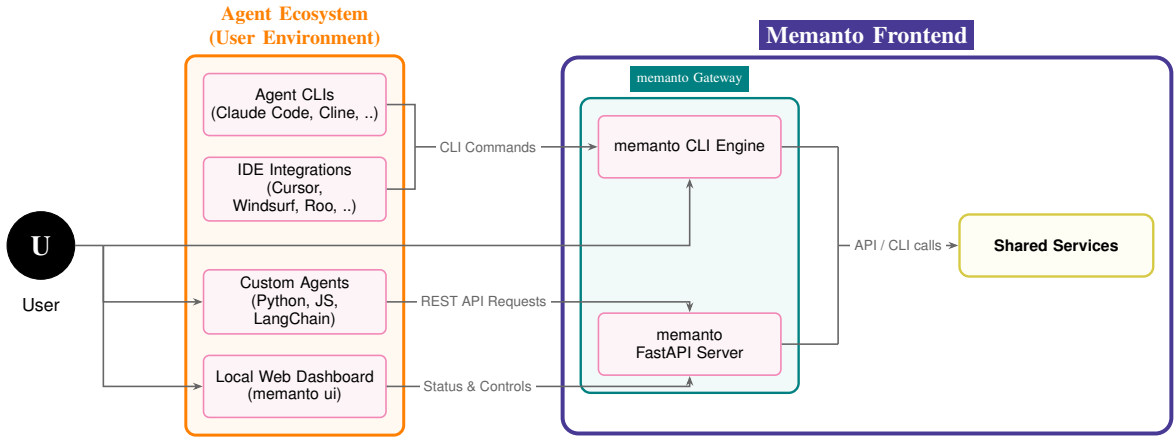
\begin{figure*}[!h]
\centering
% \hspace{1.0cm}
\scalebox{.75}{
\begin{tikzpicture}[
    node distance=0.5cm and 1.2cm,
    memanto_pink/.style={draw=magenta!60, fill=magenta!5},
    usernode/.style={circle, draw=black, fill=black, minimum size=1.2cm, text=white},
    component/.style={rectangle, rounded corners=3pt, memanto_pink, line width=0.8pt,
        minimum width=3.2cm, minimum height=1.1cm, text centered, text width=3cm,
        font=\footnotesize\sffamily},
    service/.style={rectangle, rounded corners=2pt, memanto_pink, line width=0.8pt,
        minimum width=3cm, minimum height=0.7cm, text centered, text width=2.8cm,
        font=\footnotesize\sffamily},
    sharednode/.style={rectangle, rounded corners=5pt, draw=yellow!80!black, fill=yellow!5,
        line width=1.2pt, minimum width=3.4cm, minimum height=1.1cm,
        text centered, text width=3.2cm, font=\footnotesize\sffamily\bfseries},
    arr/.style={-{Stealth[length=5pt]}, thick, gray!80!black},
    line/.style={thick, gray!80!black},
    arrlab/.style={font=\scriptsize\sffamily, fill=white, inner sep=2pt},
]

%% NODES
\node (user) [usernode] at (-4.5,0) {\Large \textbf{U}};
\node [below=0.2cm of user, font=\small\sffamily] {User};

% Agent Ecosystem
\node (agcli)  [component] at (0.0,  2.5) {Agent CLIs\\(Claude Code, Cline, ..)};
\node (ide)    [component] at (0.0,  1.0) {IDE Integrations\\(Cursor, Windsurf, Roo, ..)};
\node (custom) [component] at (0.0, -1.0) {Custom Agents\\(Python, JS, LangChain)};
\node (webui)  [component] at (0.0, -2.5) {Local Web Dashboard\\(memanto ui)};

% Gateway
\node (mcli)  [component] at (7.0,  1.75) {memanto CLI Engine};
\node (mserv) [component] at (7.0, -1.75) {memanto FastAPI Server};

% Shared Services summary node — yellow to match backend
\node (shared) [sharednode] at (13.5, 0.0) {Shared Services};

%% CONTAINERS
\begin{scope}[on background layer]
    \node[draw=orange, line width=1.5pt, rounded corners=5pt, fill=orange!5,
          fit=(agcli)(webui), inner sep=0.3cm,
          label={[orange, font=\bfseries, align=center]north:{Agent Ecosystem\\(User Environment)}}]{};
    \node[draw=teal, line width=1.2pt, rounded corners=5pt, fill=teal!5,
          fit=(mcli)(mserv), inner sep=0.3cm,
          label={[white, fill=teal, font=\scriptsize, yshift=2.5]north:memanto Gateway}] (gw) {};
    \node[draw=BlueViolet, line width=2pt, rounded corners=8pt,
          fit=(gw)(shared), inner xsep=0.3cm, inner ysep=0.7cm,
          label={[white, fill=BlueViolet, font=\large\bfseries, yshift=2.5]north:Memanto Frontend}] {};
\end{scope}

%% CONNECTIONS — User → Agent Ecosystem
\draw[line] (user.east) -- ++(0.5,0) coordinate (u_fork);
\draw[arr]  (u_fork) |- (custom.west);
\draw[arr]  (u_fork) |- (webui.west);
\draw[arr]  (u_fork) -| (mcli.south);

\draw[line] (ide.east)   -| ++(.5,0.75) coordinate (cli_merge);
\draw[line] (agcli.east) -| (cli_merge);
\draw[arr]  (cli_merge) -- node[arrlab, pos=.40]{CLI Commands} (mcli.west);

\draw[arr] (custom.east) -| node[arrlab, pos=0.16]{REST API Requests} (mserv.north);

\draw[line] (webui.east) -- ++(0.0,0) coordinate (ui_merge);
\draw[arr]  (ui_merge) -| node[arrlab, pos=.15]{Status \& Controls} (mserv.south);

%% CONNECTIONS — Gateway → Shared Services
\draw[line] (mcli.east)  -- ++(1,0) coordinate (b1);
\draw[line] (mserv.east) -- ++(1,0) coordinate (b2);
\draw[line] (b1) -- (b2);
\draw[arr]  ($(b1)!0.5!(b2)$) -- node[arrlab]{API / CLI calls} (shared.west);

\end{tikzpicture}
}
\caption{Memanto Frontend Architecture: User, Agent Ecosystem, and memanto Gateway}
\label{fig:memanto-frontend}
\end{figure*}

Fig. \ref{fig:memanto-backend} presents the Memanto Backend Architecture, which also comprises two layers. \textbf{(1) Shared Services.} The gateway routes requests to nine internal services: Daily Summary, Conflict Resolution, Answer, Recall, Remember, Agent Manager, Session and Authentication Manager, Memory Sync via \textsc{memory.md} injection, and Tool Connect. The first six services communicate directly with the Moorcheh cloud layer via SDK calls, while the remaining three operate internally. \textbf{(2) Moorcheh.ai Cloud Layer.} All storage and retrieval pass through the Moorcheh Engine API to three components: a zero-indexing semantic database, an agent-optimized RAG pipeline, and native LLM access.

\begin{figure*}[!h]
% \hspace{2 cm}
\centering
\scalebox{.8}{
\begin{tikzpicture}[
    node distance=0.5cm and 1.2cm,
    memanto_pink/.style={draw=magenta!60, fill=magenta!5},
    component/.style={rectangle, rounded corners=3pt, memanto_pink, line width=0.8pt,
        minimum width=3.2cm, minimum height=1.1cm, text centered, text width=3cm,
        font=\footnotesize\sffamily},
    service/.style={rectangle, rounded corners=2pt, memanto_pink, line width=0.8pt,
        minimum width=3cm, minimum height=0.7cm, text centered, text width=2.8cm,
        font=\footnotesize\sffamily},
    arr/.style={-{Stealth[length=5pt]}, thick, gray!80!black},
    line/.style={thick, gray!80!black},
    arrlab/.style={font=\scriptsize\sffamily, fill=white, inner sep=2pt},
]

%% SHARED SERVICES — 6 SDK-connected
\node (daily)    [service] at (3.5,  4.0) {Daily Summary};
\node (conflict) [service] at (3.5,  3.0) {Conflict Resolution};
\node (answer)   [service] at (3.5,  2.0) {Answer};
\node (recall)   [service] at (3.5,  1.0) {Recall};
\node (remembr)  [service] at (3.5,  0.0) {Remember};
\node (amgr)     [service] at (3.5, -1.0) {Agent Manager};

%% SHARED SERVICES — 3 internal-only
\node (session)  [service] at (3.5, -2.0) {Session \& Auth Manager};
\node (memsync)  [service] at (3.5, -3.0) {Memory Sync\\(MEMORY.md inject)};
\node (toolcon)  [service] at (3.5, -4.0) {Tool Connect};

%% MOORCHEH.AI components
\node (engapi) [component] at (10.0,  0.0) {Moorcheh Engine API};
\node (rag)    [component] at (15.5,  0.0) {Agentic Optimized RAG};
\node (zerodb) [component] at (15.5,  1.5) {Zero-Indexing Semantic Database};
\node (llmnat) [component] at (15.5, -1.5) {Native Access to LLM};

%% CONTAINERS
\begin{scope}[on background layer]
    \node[draw=yellow!80!black, line width=1.2pt, rounded corners=5pt, fill=yellow!5,
          fit=(daily)(toolcon), inner sep=0.3cm,
          label={[black, fill=yellow, font=\scriptsize, yshift=2.5]north:Shared Services}] (shared) {};
    \node[draw=violet!70, line width=1.5pt, rounded corners=5pt, fill=violet!5,
          fit=(engapi)(zerodb)(llmnat), inner sep=0.3cm,
          label={[white, fill=violet, font=\bfseries, yshift=2.5]north:Moorcheh.ai}] (moorcheh) {};
    \node[draw=BlueViolet, line width=2pt, rounded corners=8pt,
          fit=(shared)(moorcheh), inner xsep=0.3cm, inner ysep=0.7cm,
          label={[white, fill=BlueViolet, font=\large\bfseries, yshift=-20.5]north:Memanto Backend}] {};
\end{scope}

%% CONNECTIONS — 6 SDK services → Moorcheh Engine API
\draw[line] (daily.east)    -- ++(0.2,0) coordinate (r1);
\draw[line] (conflict.east) -- ++(0.2,0) coordinate (r2);
\draw[line] (answer.east)   -- ++(0.2,0) coordinate (r3);
\draw[line] (recall.east)   -- ++(0.2,0) coordinate (r4);
\draw[line] (remembr.east)  -- ++(0.2,0) coordinate (r5);
\draw[line] (amgr.east)     -- ++(0.2,0) coordinate (r6);
\draw[line] (r1) -- (r6);
\draw[arr]  ($(r1)!0.5!(r6)$) |- node[arrlab, pos=.76]{SDK Calls} (engapi.west);

%% CONNECTIONS — Moorcheh Engine API → internal components
\draw[line] (engapi.east) -- ++(0.1,0) coordinate (m_bus);
\draw[arr]  (m_bus) |- (zerodb.west);
\draw[arr]  (m_bus) --     (rag.west);
\draw[arr]  (m_bus) |- (llmnat.west);

\end{tikzpicture}
}
\caption{Memanto Backend Architecture: Shared Services and Moorcheh.ai Integration}
\label{fig:memanto-backend}
\end{figure*}
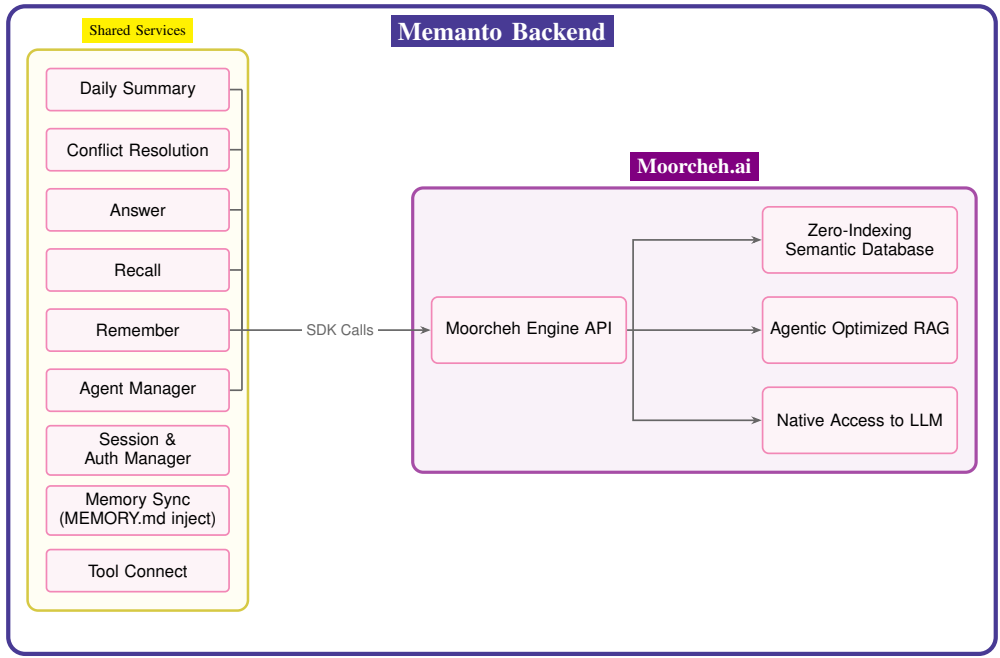

\subsection{The Moorcheh Foundation: Information Theoretic Search}

Memanto's retrieval capabilities are built on Moorcheh's semantic engine, which departs fundamentally from the HNSW plus cosine distance paradigm that dominates traditional vector databases. The engine employs three core algorithmic innovations:

\noindent\textbf{Maximally Informative Binarization (MIB):} Compresses high-dimensional floating-point embedding vectors into compact binary representations while preserving the information-theoretic content relevant to retrieval. This achieves 32$\times$ compression with no measurable loss in retrieval-relevant signal.

\noindent\textbf{Efficient Distance Metric (EDM):} Replaces cosine similarity with an information-theoretic distance measure that scores memory chunks by their ability to reduce uncertainty in the query context, rather than by surface-level geometric proximity in embedding space.

\noindent\textbf{Information Theoretic Score (ITS):} A universal relevance score on a normalized $[0,1]$ scale that quantifies the decision-theoretic value of each retrieved chunk for the current query. ITS enables deterministic, threshold-based retrieval: the same query always produces the same results, a critical property for reproducible agent behavior in regulated environments.

Together, these innovations eliminate the need for index construction, enabling instant write-to-search availability. The Moorcheh engine has been independently validated on the MAIR (Massive AI Retrieval) benchmark, achieving 64--74\% NDCG@10 with 9.6\,ms distance calculation latency (compared to 37--86\,ms for PGVector and Qdrant), sustaining 2,000+ queries per second with zero accuracy degradation, and delivering a 6.6$\times$ end-to-end speedup versus Pinecone plus Cohere reranking pipelines \cite{abtahi2025hnswinformationtheoreticbinarizationrethinking}.

\begin{table*}[htbp]
\centering
\caption{Memanto typed memory schema with 13 semantic categories.}
\label{tab:schema}
\small
\begin{tabular}{@{}L{1.8cm}L{3.5cm}L{3.5cm}L{2.5cm}@{}}
\toprule
\textbf{Type} & \textbf{Description} & \textbf{Example} & \textbf{Priority Signal} \\
\midrule
\memtype{fact}         & Objective, verifiable information   & User is in PST timezone            & Stable, high confidence \\
\rowcolor{lightgray}
\memtype{preference}   & User or system preferences          & Prefers dark mode                  & Moderate decay \\
\memtype{decision}     & Choices made affecting future       & Chose PostgreSQL for DB            & High persistence \\
\rowcolor{lightgray}
\memtype{commitment}   & Promises or obligations             & Deliver report by Friday           & Time critical \\
\memtype{goal}         & Objectives to achieve               & Reach 10K users by Q4              & Active until achieved \\
\rowcolor{lightgray}
\memtype{event}        & Historical occurrences              & Meeting with CEO at 2pm            & Episodic, decaying \\
\memtype{instruction}  & Rules and guidelines                & Always validate input              & Procedural, persistent \\
\rowcolor{lightgray}
\memtype{relationship} & Entity connections                  & Alice manages Bob                  & Graph-like, stable \\
\memtype{context}      & Situational information             & Currently in budget review         & Highly temporal \\
\rowcolor{lightgray}
\memtype{learning}     & Lessons from experience             & Users need simpler onboarding      & Accumulating \\
\memtype{observation}  & Patterns noticed                    & Traffic peaks on Fridays           & Statistical, evolving \\
\rowcolor{lightgray}
\memtype{error}        & Mistakes to avoid                   & Do not use deprecated API          & Persistent guard \\
\memtype{artifact}     & Document or code references         & Q3 budget spreadsheet              & Reference pointer \\
\bottomrule
\end{tabular}
\end{table*}

\subsection{Typed Memory Schema}

Memanto implements a typed memory schema with 13 built-in semantic categories, each carrying distinct retrieval semantics and priority weighting. This design is directly motivated by the cognitive science literature on memory type distinctions \cite{tulving1972episodic} and the empirical findings of ENGRAM \cite{patel2025engram} that typed separation significantly outperforms undifferentiated storage.
A comprehensive list of all available memory types and categories is provided in Table \ref{tab:schema}.

This typing system serves a dual purpose. First, it enables type-filtered retrieval, allowing agents to query specifically for commitments, decisions, or any other category without polluting the result set. Second, it provides implicit priority and decay signals that the retrieval engine uses to weight results appropriately.

\subsection{Conflict Resolution}

A distinguishing feature of Memanto is its built-in conflict resolution mechanism. When a new memory is committed that semantically contradicts an existing memory, for example, \textit{Project deadline is April 15} followed by \textit{Project deadline is May 1}, Memanto detects the conflict and notifies the agent, requesting explicit resolution before the contradiction is persisted.

This mechanism directly addresses the \emph{constraint drift} failure mode identified in long-running agent deployments. Without conflict detection, agents gradually accumulate contradictory memories that erode the coherence of their reasoning. MemoryAgentBench \cite{memoryagentbench2025} confirms that conflict resolution remains one of the most significant unsolved challenges in current memory systems, with all evaluated methods failing on multi-hop conflict scenarios.

Conflict resolution operates through semantic similarity matching against existing memories of the same type within the agent's namespace, using a configurable contradiction threshold. When triggered, the system presents the agent with the conflicting memories and three resolution options: \emph{supersede} (replace the old memory), \emph{retain} (keep the old memory), or \emph{annotate} (preserve both with a conflict flag for human review).

\subsection{Temporal Versioning}

Memanto supports three temporal query modalities:

\begin{itemize}
    \item \textbf{As-of queries:} Retrieve the memory state as it existed at a specific historical timestamp, enabling audit trail reconstruction.
    \item \textbf{Changed-since queries:} Retrieve all memories created or modified within a time range, supporting incremental context updates.
    \item \textbf{Current-only queries:} Retrieve only non-superseded memories, providing ground-truth state without historical noise.
\end{itemize}

Memory supersession is non-destructive: superseded entries are marked accordingly but retained in the store, enabling full temporal reconstruction. This design is critical for compliance-sensitive deployments in regulated industries and directly addresses the knowledge update evaluation category in LongMemEval \cite{wu2024longmemeval}.

\subsection{Session and Namespace Management}

Memanto isolates agent memory through a namespace architecture in which each agent maintains an independent memory namespace. Sessions are time-bounded windows with a default duration of six hours that provide temporal grouping without restricting cross-session retrieval. All memories within a namespace remain accessible regardless of session boundaries, enabling the persistent cross-session context that constitutes the primary use case for agentic memory.

\subsection{Daily Intelligence}

Memanto generates automated daily intelligence artifacts including session summaries, contradiction detection reports, and interactive conflict resolution prompts. These artifacts are persisted as local Markdown files and optionally synced to Moorcheh's cloud store, providing both human-readable audit trails and machine-queryable context for agents operating on daily planning cycles.
\section{Experimental Evaluation and Results}\label{sec:exp}

Memanto is evaluated on two established agentic memory benchmarks through a five-stage progressive ablation study in which the independent contribution of each architectural decision is systematically isolated and quantified. Per-category accuracy results are subsequently reported for the final configuration, followed by a comprehensive comparison against all publicly available competing systems and a quantitative analysis of the operational overhead imposed by alternative architectures. Memanto is publicly accessible as a Python package via \texttt{pip install memanto}.

\subsection{Benchmarks and Evaluation Protocol}

\textbf{\sys{LongMemEval}$_S$}~\cite{wu2024longmemeval}: A large-scale benchmark comprising 500 manually curated questions distributed across six categories, designed to evaluate five core memory abilities including information extraction, multi-session reasoning, temporal reasoning, knowledge update, and abstention. The standard evaluation setting uses approximately 115K tokens across approximately 50 sessions. All evaluations employ Claude Sonnet~4 as the LLM judge.

\textbf{\sys{LoCoMo}}~\cite{maharana2024locomo}: A multi-modal dialogue benchmark spanning four reasoning categories, namely Single-Hop, Multi-Hop, Open Domain, and Temporal reasoning. Individual dialogues extend to 35 sessions, 300 turns, and approximately 9K tokens on average, providing a rigorous test of long-horizon conversational memory.

To ensure evaluation consistency across systems, answer generation and judge prompts are adapted from the Hindsight repository~\cite{hindsight2025}, which is specifically designed to mitigate two systematic evaluation artifacts: answerer refusal on ambiguous questions, and rigid judge rejection of responses that are semantically correct but lexically divergent from the reference answer. All experiments employ Memanto's vector-only architecture with the Moorcheh ITS engine as the sole retrieval backend.

\subsection{Progressive Ablation Study}

Rather than reporting only final results, the contribution of each architectural decision is evaluated in isolation through a controlled, sequential ablation. Each stage is defined by a specific configuration change, and its effect is measured against the preceding stage. The five stages are presented below, each accompanied by its configuration, quantitative outcomes, and the principal empirical finding it yields.

\subsubsection{Stage 1: Naive Baseline}

\begin{table}[!h]
\caption{Stage 1: Naive Baseline}
\label{tab:s1}
\centering\small
\begin{tabular}{@{}lc@{}}
\toprule
\textbf{Benchmark} & \textbf{Accuracy} \\ \midrule
\sys{LongMemEval}  & 56.6\% \\
\sys{LoCoMo}       & 76.2\% \\
\bottomrule
\multicolumn{2}{l}{\scriptsize Config: $k$=10, threshold=0.15, Claude Sonnet 4}
\end{tabular}
\end{table}

\textit{Configuration:} Standard semantic search with a retrieval limit of $k$=10, an ITS similarity threshold of 0.15, and Claude Sonnet~4 as the inference model. As shown in Table~\ref{tab:s1}, this configuration yields 56.6\% on \sys{LongMemEval} and 76.2\% on \sys{LoCoMo}, establishing the performance floor for a minimally parameterised retrieval-augmented generation implementation on Memanto. The 19.6 percentage point gap between the two benchmarks does not reflect a difference in session length, as the two datasets are comparable in this regard. Rather, it reflects a difference in question structure and information accessibility. \sys{LongMemEval} queries tend to be longer and span multiple topics simultaneously, which distributes the semantic signal across a broader embedding space and reduces similarity scores for relevant chunks. Under a threshold of 0.15, this information frequently falls below the retrieval cutoff and is not surfaced. The gap therefore serves as an early indicator of the sensitivity of retrieval performance to threshold calibration, a dynamic that Stage~2 directly exploits.

\subsubsection{Stage 2: Recall Expansion}

Standard RAG practice constrains retrieval to a fixed top-$k$ of 10, applying aggressive precision filtering at the retrieval layer. For agentic memory, however, multi-session questions frequently require the synthesis of facts that are distributed across temporally disjoint sessions~\cite{wu2024longmemeval}, rendering this constraint a critical architectural bottleneck.

\textit{Configuration:} Retrieval limit increased to 40 chunks; ITS similarity threshold relaxed to 0.10.

As shown in Table~\ref{tab:s2}, this single parameter adjustment yields a gain of 20.4 percentage points on \sys{LongMemEval} and 6.6 percentage points on \sys{LoCoMo}, constituting the largest single improvement observed across all five ablation stages.

\begin{table}[!h]
\caption{Stage 2: Recall Expansion ($k=10 \rightarrow 40$)}
\label{tab:s2}
\centering\small
\begin{tabular}{@{}lcc@{}}
\toprule
\textbf{Benchmark} & \textbf{Accuracy} & $\boldsymbol{\Delta}$ \\ \midrule
\sys{LongMemEval}  & 77.0\% & $+$20.4 percentage points \\
\sys{LoCoMo}       & 82.8\% & $+$6.6 percentage points \\
\bottomrule
\end{tabular}
\end{table}

\noindent\textit{Finding.} The precision-recall tradeoff in agentic memory skews decisively toward recall. Providing the LLM with a broader, noisier candidate set and relying on its in-context reasoning to filter irrelevant content is substantially more effective than constraining retrieval to a narrow, high-precision window that risks excluding critical fragments.

\subsubsection{Stage 3: Prompt Optimization}

\textit{Configuration:} Retrieval parameters unchanged from Stage~2 ($k$=40, threshold=0.10); answer generation and judge prompts replaced with optimized variants from the Hindsight repository~\cite{hindsight2025}.

As shown in Table~\ref{tab:s3}, this modification yields marginal improvements of 2.2 percentage points on \sys{LongMemEval} and 0.1 percentage points on \sys{LoCoMo}.

\begin{table}[!ht]
\caption{Stage 3: Prompt Optimization}
\label{tab:s3}
\centering\small
\begin{tabular}{@{}lcc@{}}
\toprule
\textbf{Benchmark} & \textbf{Accuracy} & $\boldsymbol{\Delta}$ \\ \midrule
\sys{LongMemEval}  & 79.2\% & $+$2.2 percentage points \\
\sys{LoCoMo}       & 82.9\% & $+$0.1 percentage points \\
\bottomrule
\end{tabular}
\end{table}

\noindent\textit{Finding.} Prompt engineering yields only marginal performance improvements. When the retrieval layer fails to surface relevant content, no degree of prompt refinement compensates for the resulting structural deficit. This finding is consistent with the broader observation that as foundational model capabilities continue to advance, the relative contribution of prompt design to overall system performance diminishes in proportion to the quality of the underlying retrieval mechanism.

\subsubsection{Stage 4: Maximum Recall}

Error analysis of Stage~3 failures established that incorrect answers were attributable not to LLM confusion arising from expanded context~\cite{liu2024lost}, but to semantic search consistently failing to retrieve critical sentences embedded within otherwise largely irrelevant chunks.

\textit{Configuration:} Dynamic retrieval limit expanded to a maximum of 100 chunks, governed by ITS-threshold gating rather than a fixed-$k$ constraint; similarity threshold lowered to 0.05.

As shown in Table~\ref{tab:s4}, this configuration yields improvements of 5.8 percentage points on \sys{LongMemEval} and 3.4 percentage points on \sys{LoCoMo}.

\begin{table}[!h]
\caption{Stage 4: Maximum Recall ($k=100$)}
\label{tab:s4}
\centering\small
\begin{tabular}{@{}lcc@{}}
\toprule
\textbf{Benchmark} & \textbf{Accuracy} & $\boldsymbol{\Delta}$ \\ \midrule
\sys{LongMemEval}  & 85.0\% & $+$5.8 percentage points \\
\sys{LoCoMo}       & 86.3\% & $+$3.4 percentage points \\
\bottomrule
\end{tabular}
\end{table}

\noindent\textit{Finding.} Modern large language models exhibit a high degree of tolerance for noisy retrieval context. High-dimensional vector search is susceptible to distortion by multi-topic chunks in which a single critical detail is co-located with predominantly irrelevant content. Extending the retrieval budget to accommodate such cases consistently outperforms engineering for retrieval precision, confirming that recall is the dominant lever for agentic memory performance.

\subsubsection{Stage 5: Inference Model Upgrade}

\textit{Configuration:} Inference model upgraded from Claude Sonnet~4 to Gemini~3, establishing parity with the leading competing systems against which Memanto is benchmarked. This stage isolates the contribution of inference model capability from that of the memory architecture, ensuring that comparative results reflect architectural differences rather than model selection artifacts.

\begin{figure}[!h]
\centering
\scalebox{.83}{
\begin{tikzpicture}
\begin{axis}[
  xbar, bar width=7pt,
  width=\columnwidth, height=6.2cm,
  xmin=50, xmax=101,
  xlabel={Accuracy (\%)},
  xlabel style={font=\small},
  symbolic y coords={
    {S1 Baseline},
    {S2 Recall},
    {S3 Prompt},
    {S4 MaxRecall},
    {S5 Model}},
  ytick=data,
  yticklabel style={font=\small, rotate= 0},
  enlarge y limits=0.18,
  nodes near coords,
  nodes near coords style={font=\tiny,anchor=west},
  nodes near coords align={horizontal},
  legend style={at={(1.,0.02)},anchor=south east,font=\small},
  ymajorgrids, grid style={dashed,gray!20},
]
\addplot[fill=navyblue!80,draw=navyblue] coordinates {
  (56.6,{S1 Baseline})
  (77.0,{S2 Recall})
  (79.2,{S3 Prompt})
  (85.0,{S4 MaxRecall})
  (89.8,{S5 Model})};
\addlegendentry{\sys{LongMemEval}}
\addplot[fill=medblue!45,draw=medblue] coordinates {
  (76.2,{S1 Baseline})
  (82.8,{S2 Recall})
  (82.9,{S3 Prompt})
  (86.3,{S4 MaxRecall})
  (87.1,{S5 Model})};
\addlegendentry{\sys{LoCoMo}}
\end{axis}
\end{tikzpicture}
}
\caption{Progressive ablation waterfall. Stage 2 (recall expansion, $k=10 \rightarrow 40$) delivers the largest single gain (+20.4 percentage points on \sys{LongMemEval}), confirming that retrieval tuning rather than architectural complexity is the dominant performance driver. Stage 3 (prompt optimisation) contributes only +2.2 percentage points, while Stages 4 and 5 contribute +5.8 and +4.8 percentage points respectively.}
\label{fig:ablation}
\end{figure}
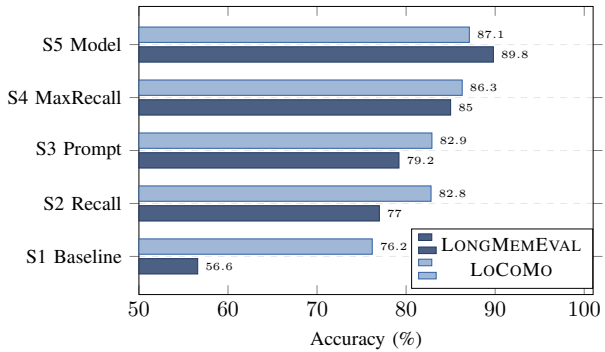

The complete ablation progression across all five stages, together with the cumulative accuracy gains at each step, is visualized in Fig. \ref{fig:ablation}. The relationship between retrieval limit $k$ and accuracy, overlaid with average token cost per query, is presented in Fig. \ref{fig:kplot}.

\begin{figure}[!h]
\centering
\scalebox{.85}{
\begin{tikzpicture}
\begin{axis}[
  name=mainax,
  width=\columnwidth, height=5.6cm,
  xlabel={Retrieval limit $k$},
  ylabel={Accuracy (\%)},
  xlabel style={font=\small},
  ylabel style={font=\small},
  xmin=5, xmax=108,
  ymin=50, ymax=95,
  xtick={10,20,40,60,80,100},
  ymajorgrids, xmajorgrids,
  grid style={dashed,gray!22},
  legend pos=south east,
  legend style={font=\small},
  axis y line*=left,
  tick label style={font=\small},
]
\addplot[thick,navyblue,mark=*,mark size=2pt]
  coordinates{(10,56.6)(20,67.1)(40,77.0)(60,80.5)(80,83.2)(100,85.0)};
\addlegendentry{\sys{LME}}
\addplot[thick,medblue,dashed,mark=square*,mark size=1.8pt]
  coordinates{(10,76.2)(20,79.5)(40,82.8)(60,84.4)(80,85.5)(100,86.3)};
\addlegendentry{\sys{LoCoMo}}
\draw[colorHyb,dashed,thin](40,50)--(40,82.8);
\node[font=\tiny,colorHyb,anchor=south west] at(41,83.2){Inflection $k=40$};
\end{axis}

\begin{axis}[
  at={(mainax.south west)},
  anchor=south west,
  width=\columnwidth, height=5.6cm,
  xmin=5, xmax=108,
  ymin=0, ymax=52000,
  axis y line*=right,
  axis x line=none,
  ylabel={Avg. tokens / query},
  ylabel style={font=\small},
  ytick={10000,20000,30000,40000,50000},
  yticklabels={10K,20K,30K,40K,50K},
  tick label style={font=\small},
]
\addplot[colorHyb,dotted,thick,mark=triangle,mark size=1.5pt]
  coordinates{(10,4200)(20,8800)(40,17500)(60,26000)(80,35000)(100,42000)};
\end{axis}
\end{tikzpicture}
}
\caption{Accuracy versus retrieval limit $k$ (left axis) and average tokens consumed per query (right axis, dotted). Both accuracy curves plateau above $k=60$, with a clear inflection at $k=40$. The accuracy gain from $k=10 \rightarrow 40$ (+20.4 percentage points on \sys{LME}) far outweighs the approximately fourfold token cost increase, validating the recall over precision principle.}
\label{fig:kplot}
\end{figure}
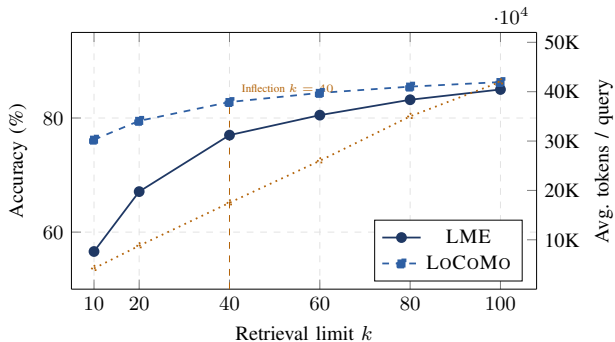

\subsection{Final Results by Category}

Per-category accuracy results at Stage~5 under Gemini~3 inference are reported in Tables~\ref{tab:lme_final} and~\ref{tab:locomo_final} for \sys{LongMemEval} and \sys{LoCoMo}, respectively. Memanto achieves 100.0\% on the Single-session Assistant category and 95.7\% on Single-session User, with the lowest performance on Multi-session queries at 81.2\%, reflecting the inherent difficulty of synthesizing information distributed across extended interaction histories.

\begin{table}[!h]
\caption{\sys{LongMemEval} Final Results by Category (Stage~5, Gemini~3)}
\label{tab:lme_final}
\centering\small
\begin{tabular}{@{}lc@{}}
\toprule
\textbf{Category} & \textbf{Accuracy} \\ \midrule
Single-session User        & 95.7\%  \\
\rowcolor{rowgray}
Single-session Assistant   & 100.0\% \\
Single-session Preference  & 93.3\%  \\
\rowcolor{rowgray}
Knowledge Update           & 93.6\%  \\
Temporal Reasoning         & 88.0\%  \\
\rowcolor{rowgray}
Multi-session              & 81.2\%  \\
\midrule
\textbf{Overall}           & \textbf{89.8\%} \\
\bottomrule
\end{tabular}
\end{table}

\begin{table}[!h]
\caption{\sys{LoCoMo} Final Results by Category (Stage~5, Gemini~3)}
\label{tab:locomo_final}
\centering\small
\begin{tabular}{@{}lc@{}}
\toprule
\textbf{Category} & \textbf{Accuracy} \\ \midrule
Single-Hop  & 78.7\% \\
\rowcolor{rowgray}
Multi-Hop   & 70.8\% \\
Open Domain & 92.4\% \\
\rowcolor{rowgray}
Temporal    & 85.4\% \\
\midrule
\textbf{Overall} & \textbf{87.1\%} \\
\bottomrule
\end{tabular}
\end{table}

\begin{table*}[!ht]
\caption{System Comparison on \sys{LongMemEval} and \sys{LoCoMo}}
\label{tab:comparison}
\centering\small
\scalebox{.9}{
\begin{tabular}{@{}L{2.8cm}C{1.3cm}C{1.3cm}L{3.2cm}L{1.5cm}L{2.0cm}@{}}
\toprule
\textbf{System} & \textbf{LoCoMo} & \textbf{LMEval}
  & \textbf{Architecture} & \textbf{Retrieval} & \textbf{Query Strategy} \\
\midrule
\rowcolor{lightblue}
\textbf{Memanto (ours)}
  & \textbf{87.1\%} & \textbf{89.8\%}
  & Vector Only & RAG & Single Query \\
Hindsight~\cite{hindsight2025}
  & 89.6\% & 91.4\%
  & Hybrid (Reflection and Vector) & Parallel & Multi-Query \\
\rowcolor{rowgray}
EmergenceMem
  & --- & 86.0\%
  & Hybrid (Graph and Vector) & Parallel & Multi-Query \\
Supermemory
  & --- & 85.2\%
  & Hybrid (Graph and Vector) & Parallel & Multi-Query \\
\rowcolor{rowgray}
% ENGRAM~\cite{patel2025engram}
%   & ${\approx}$80\% & ${\approx}$78\%
%   & Vector (Typed) & Top-$k$ & Single Query \\
Memobase
  & 75.8\% & ---
  & Hybrid (Graph and Vector) & Parallel & Single Query \\
\rowcolor{rowgray}
Zep~\cite{rasmussen2025zep}
  & 75.1\% & 71.2\%
  & Hybrid (Graph and Vector) & Parallel & Single Query \\
Letta~\cite{packer2024memgptllmsoperatingsystems}
  & 74.0\% & ---
  & Local Filesystem & RAG & Recursive \\
\rowcolor{rowgray}
Full Context
  & 72.9\% & 60.2\%
  & Full Context & Full & Not applicable \\
Mem0$_g$~\cite{chhikara2025mem0buildingproductionreadyai}
  & 68.4\% & ---
  & Hybrid (Graph and Vector) & Parallel & Single Query \\
\rowcolor{rowgray}
Mem0~\cite{chhikara2025mem0buildingproductionreadyai}
  & 66.9\% & ---
  & Vector Only & Parallel & Single Query \\
LangMem
  & 58.1\% & ---
  & Vector Only & RAG & Single Query \\
\bottomrule
\end{tabular}
}
\end{table*}

\subsection{Comparative Results}

Table~\ref{tab:comparison} presents a comprehensive comparison of Memanto against all publicly reported results from competing agentic memory systems on both benchmarks. All figures for competing systems are drawn from their respective published papers or official benchmark reports. As shown in Table~\ref{tab:comparison}, Memanto achieves the highest accuracy among all vector-only systems on both benchmarks, surpassing Mem0 by 22.9 percentage points on \sys{LongMemEval} and 20.2 percentage points on \sys{LoCoMo}. Hindsight attains higher overall accuracy on both benchmarks, but does so at a complexity score of 4 out of 4, requiring dynamic multi-query retrieval and structured reflection passes, as visualized in Fig.~\ref{fig:scatter}. The architectural complexity score is computed as the sum of four binary indicators: (1) requires a graph database, (2) invokes an LLM at ingestion time, (3) employs multi-query retrieval, and (4) uses recursive or reflective querying. Each indicator contributes one point, yielding a scale from 0 (minimal overhead) to 4 (maximum complexity). The grouped accuracy comparison across all systems is presented in Fig.~\ref{fig:barcomp}.

\begin{figure}[!h]
\centering
\scalebox{.85}{
\begin{tikzpicture}
\begin{axis}[
  xbar, bar width=5.5pt,
  width=\columnwidth, height=8.5cm,
  xmin=50, xmax=103,
  xlabel={Accuracy (\%)}, xlabel style={font=\small},
  symbolic y coords={
    LangMem, Mem0, {Mem0$_g$}, {Full Context}, Letta, Zep,
    Memobase, ENGRAM, Supermem., EmergMem, Hindsight, Memanto},
  ytick=data, yticklabel style={font=\small},
  enlarge y limits=0.055,
  nodes near coords,
  nodes near coords style={font=\tiny,anchor=west},
  legend style={at={(0.98,0.02)},anchor=south east,font=\small},
  ymajorgrids, grid style={dashed,gray!20},
]
\addplot[fill=navyblue!80,draw=navyblue] coordinates {
  (58.1,LangMem)(66.9,Mem0)(68.4,{Mem0$_g$})(60.2,{Full Context})
  (74.0,Letta)(71.2,Zep)(75.8,Memobase)(78.0,ENGRAM)
  (85.2,Supermem.)(86.0,EmergMem)(91.4,Hindsight)(89.8,Memanto)};
\addlegendentry{\sys{LongMemEval}}
\addplot[fill=medblue!40,draw=medblue] coordinates {
  (58.1,LangMem)(66.9,Mem0)(68.4,{Mem0$_g$})(72.9,{Full Context})
  (74.0,Letta)(75.1,Zep)(75.8,Memobase)(80.0,ENGRAM)
  (75.8,Supermem.)(80.0,EmergMem)(89.6,Hindsight)(87.1,Memanto)};
\addlegendentry{\sys{LoCoMo}}
\end{axis}
\end{tikzpicture}
}
\caption{Benchmark comparison. Memanto leads vector-only systems.
  Hindsight achieves higher accuracy via multi-query parallel
  retrieval + reflection layers (complexity score 4/4).}
\label{fig:barcomp}
\end{figure}
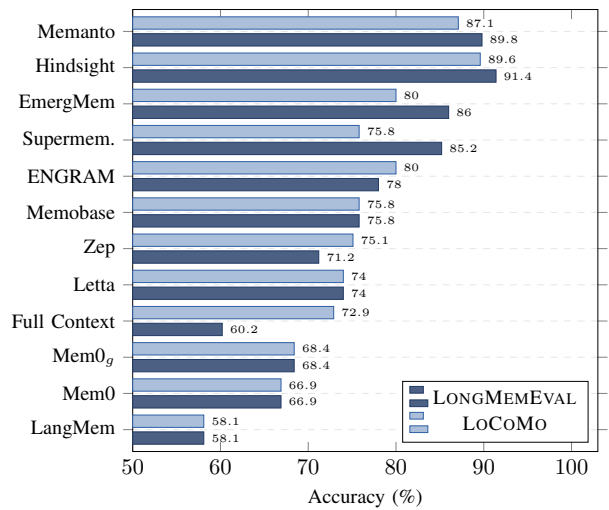
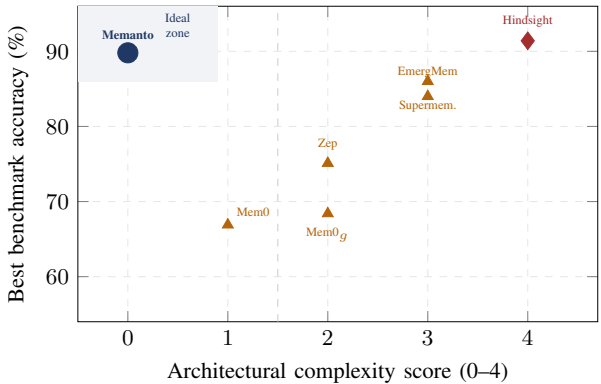
\begin{figure}[!ht]
\centering
\scalebox{.95}{
\begin{tikzpicture}
\begin{axis}[
  width=\columnwidth, height=6.0cm,
  xlabel={Architectural complexity score (0--4)},
  ylabel={Best benchmark accuracy (\%)},
  xlabel style={font=\small}, ylabel style={font=\small},
  xmin=-0.5, xmax=4.7, ymin=54, ymax=96,
  xtick={0,1,2,3,4},
  ymajorgrids, xmajorgrids, grid style={dashed,gray!20},
  tick label style={font=\small}, clip=false,
]
\fill[navyblue!06](-0.5,86) rectangle (0.9,96);
\node[font=\tiny,navyblue,align=center]at(0.5,93.5){Ideal\\zone};
\draw[gray!40,dashed,thin](1.5,54)--(1.5,96);
\addplot[only marks,mark=*,mark size=4pt,navyblue,
  nodes near coords={\tiny\textbf{Memanto}},
  every node near coord/.style={anchor=south,navyblue,
    font=\tiny,yshift=2pt}]
  coordinates{(0,89.8)};
% \addplot[only marks,mark=square*,mark size=2.5pt,colorVec,
%   nodes near coords={\tiny ENGRAM},
%   every node near coord/.style={anchor=south,colorVec,
%     font=\tiny,yshift=2pt}]
%   coordinates{(0,80.0)};
\addplot[only marks,mark=triangle*,mark size=2.5pt,colorHyb,
  nodes near coords={\tiny Mem0},
  every node near coord/.style={anchor=south west,colorHyb,
    font=\tiny}]
  coordinates{(1,66.9)};
\addplot[only marks,mark=triangle*,mark size=2.5pt,colorHyb,
  nodes near coords={\tiny Zep},
  every node near coord/.style={anchor=south,colorHyb,
    font=\tiny,yshift=2pt}]
  coordinates{(2,75.1)};
\addplot[only marks,mark=triangle*,mark size=2.5pt,colorHyb,
  nodes near coords={\tiny Mem0$_g$},
  every node near coord/.style={anchor=north,colorHyb,
    font=\tiny,yshift=-2pt}]
  coordinates{(2,68.4)};
\addplot[only marks,mark=triangle*,mark size=2.5pt,colorHyb,
  nodes near coords={\tiny Supermem.},
  every node near coord/.style={anchor=north,colorHyb,
    font=\tiny,yshift=2pt}]
  coordinates{(3,84)};
\addplot[only marks,mark=triangle*,mark size=2.5pt,colorHyb,
  nodes near coords={\tiny EmergMem},
  every node near coord/.style={anchor=south,colorHyb,
    font=\tiny,yshift=-2pt}]
  coordinates{(3,86.0)};
\addplot[only marks,mark=diamond*,mark size=3.5pt,memred,
  nodes near coords={\tiny Hindsight},
  every node near coord/.style={anchor=south,memred,
    font=\tiny,yshift=2pt}]
  coordinates{(4,91.4)};
\end{axis}
\end{tikzpicture}
}
\caption{Architectural complexity vs.\ accuracy. Complexity
  score $=$ sum of: requires graph~DB, LLM at ingestion,
  multi-query retrieval, recursive querying (each 0/1).
  Memanto occupies the ideal upper-left quadrant.
  Hindsight achieves higher accuracy only at maximum
  complexity~(score~4).}
\label{fig:scatter}
\end{figure}

\subsection{The Memory Tax: Overhead Analysis}

Beyond raw accuracy, sustainable production deployment requires careful consideration of the operational overhead imposed by each memory architecture. The Memory Tax is characterized along four dimensions, and its quantitative impact is summarised in Table~\ref{tab:memorytax}.

\begin{table}[!t]
\caption{Memory Tax: Operational Overhead Comparison}
\label{tab:memorytax}
\centering\small\setlength{\tabcolsep}{2.5pt}
\scalebox{.9}{
\begin{tabular}{@{}L{1.72cm}C{0.88cm}C{0.78cm}L{1.4cm}C{1.18cm}C{0.68cm}@{}}
\toprule
\textbf{System}
  & \textbf{LLM per Write}
  & \textbf{LLM per Ret.}
  & \textbf{Infrast-ructure}
  & \textbf{Ingest Latency}
  & \textbf{Idle Cost} \\
\midrule
\rowcolor{lightblue}
\textbf{Memanto}
  & 0 & 1 & Moorcheh Vector DB + API key & $<$10\,ms & Zero \\
Mem0
  & 1 & 1 & Vector DB & ${\approx}$500\,ms & Fixed \\
\rowcolor{rowgray}
Mem0$_g$
  & ${\geq}$2 & ${\geq}$2 & Vector and Neo4j
  & ${\approx}$2\,s & Fixed \\
Zep
  & ${\geq}$2 & ${\geq}$2 & Vector and Graph
  & ${\approx}$3\,s & Fixed \\
\rowcolor{rowgray}
% A-MEM
%   & ${\geq}$1 & 1 & Vector DB & ${\approx}$800\,ms & Fixed \\
\bottomrule
\end{tabular}
}
\end{table}

\textbf{Ingestion overhead.} Systems requiring LLM-mediated entity extraction, including Mem0$_g$, Zep, and A-MEM, consume tokens and incur non-trivial latency at every write operation. As shown in Table~\ref{tab:memorytax}, Mem0$_g$ and Zep invoke two or more LLM calls per write, resulting in ingestion latencies of approximately 2 and 3 seconds, respectively. For a customer support agent processing 1,000 messages per day, this overhead accumulates to a substantial operational cost. Memanto ingests raw conversational content with zero LLM invocations at write time, eliminating this cost category entirely.

\textbf{Retrieval latency.} Multi-query and recursive retrieval strategies multiply inference calls per user interaction, compounding end-to-end latency. As shown in Table~\ref{tab:memorytax}, Memanto achieves sub-10\,ms ingestion and sub-90\,ms retrieval using a single query, compared to the multi-second round-trip characteristic of graph-traversal systems.

\textbf{Infrastructure complexity.} Hybrid systems require the independent provisioning, scaling, monitoring, and maintenance of separate vector and graph database instances. As shown in Table~\ref{tab:memorytax}, Memanto requires only the Moorcheh Vector DB and API key, with no additional infrastructure to configure or operate.

\textbf{Idle cost.} Traditional vector databases mandate continuously provisioned compute resources regardless of query volume. As shown in Table~\ref{tab:memorytax}, Moorcheh's serverless architecture scales to zero during idle periods, eliminating fixed infrastructure costs for intermittent agent workloads, in contrast to the fixed idle costs incurred by all competing systems.

For an agent executing 10K daily memory operations, estimated daily costs amount to \$0.50 for Memanto, \$2.32 for Mem0-Graph, and \$1.70 for Zep, yielding annual savings of approximately \$662 per agent relative to Mem0-Graph. This difference compounds substantially across enterprise deployments with large agent fleets.

\section{Discussion}
\label{sec:discussion}

\subsection{Why Optimized Retrieval Outperforms Graph Complexity}

Our results challenge the prevailing industry consensus that knowledge graphs are necessary for high-quality agentic memory. Three factors explain Memanto's strong performance, visible in Fig. \ref{fig:scatter}'s upper-left quadrant.

\textbf{Factor 1: Retrieval reasoning decomposition.} Modern LLMs are exceptionally capable in-context reasoners when provided with relevant raw context \cite{wei2022chain}. Graph-based systems attempt to pre compute reasoning pathways through entity relationship structures, but this pre computation is inherently lossy because it must commit to a schema at index time and is therefore schema dependent. By contrast, providing the LLM a broader set of semantically relevant raw chunks and relying on its in context reasoning produces more flexible and accurate answers. Mem0's own ablation \cite{chhikara2025mem0buildingproductionreadyai} showing $\approx$2\% graph gain directly supports this interpretation.

\textbf{Factor 2: Semantic matching quality matters more than structure.} Moorcheh's ITS engine provides deterministic, exact match semantic search \cite{abtahi2025hnswinformationtheoreticbinarizationrethinking} rather than the approximate and jitter prone results of HNSW based systems \cite{malkov2018hnsw}. When the underlying search is precise and relevance scoring is information theoretically grounded in uncertainty reduction rather than geometric proximity, the structural overhead of a knowledge graph provides only diminishing marginal returns. This is consistent with ENGRAM's finding \cite{patel2025engram} that simple dense retrieval can match complex architectures.

\textbf{Factor 3: Ingestion simplicity enables faster iteration.} Eliminating the LLM extraction step at ingestion enables sub second write to retrieval feedback loops, accelerating development and debugging of agentic workflows. The synchronous extraction pipelines in Zep and Mem0$_g$ convert memory writes into multi second blocking calls, a latency profile incompatible with tight reasoning chains.

\subsection{The Recall over Precision Principle}

Fig. \ref{fig:kplot} quantifies a counter intuitive principle: in agentic memory, recall systematically outperforms precision as a retrieval objective. Expanding $k$ from 10 to 100 produces a cumulative $+$28.4\,pp improvement on LongMemEval, while prompt optimisation contributes only $+$2.2\,pp. Systems that invest engineering effort in precise structured retrieval such as graph traversal, multi hop entity resolution, and recursive query decomposition may be solving the wrong problem. The LLM is a more capable filter than any pre computed retrieval structure, at the cost of a few extra tokens of input context. This is consistent with LongMemEval's finding \cite{wu2024longmemeval} that performance continues improving beyond 20K retrieved tokens with GPT-4o, and with Liu et al.'s \cite{liu2024lost} finding that the lost in the middle effect is a function of where information appears rather than how much is retrieved.

\subsection{Conflict Resolution as a Production Necessity}

While neither LongMemEval nor LoCoMo systematically tests contradictory memories, conflict resolution is critical for production deployments. Long running agents accumulate contradictions through user corrections, updated preferences, and evolving project contexts. Without explicit conflict detection, these produce \emph{memory poisoning}, which results in increasingly incoherent agent behaviour over time. Memanto's proactive conflict detection provides a guardrail absent from all evaluated competing systems. MemoryAgentBench \cite{memoryagentbench2025} confirms all current systems fail on multi hop conflict scenarios.

\subsection{Determinism for Agentic Stability}

LLMs and autonomous agents exhibit high sensitivity to retrieval variability, where even minor changes in retrieved context can trigger divergent reasoning paths. ANN based systems introduce non determinism through probabilistic graph traversal, meaning identical queries may return different results depending on index state. This volatility compounds in multi turn agent interactions, propagating inconsistencies through conversation history. Memanto's exhaustive search architecture provides deterministic retrieval, eliminating this source of instability. Furthermore, new documents become immediately queryable without degrading search quality or requiring batch reprocessing, contrasting with HNSW systems that must balance between stale indices and computationally expensive rebuilds.

\subsection{Limitations and Future Work}

\textbf{Benchmark scope.} Both LongMemEval and LoCoMo target conversational settings. Non-conversational agentic workflows, such as research agents, code generation, and multi-agent coordination, remain untested. Merrill et al. \cite{merrill2026evaluating} identify the need for benchmark testing high-level memory organization beyond factual recall.

\textbf{Benchmark saturation and label quality.} Manual inspection of individual questions suggests that approximately 5\% of \sys{LongMemEval} questions and 6--7\% of \sys{LoCoMo} questions contain labelling inconsistencies, including ambiguous reference answers and questions whose ground truth cannot be unambiguously determined from the provided dialogue context. This label noise establishes a practical performance ceiling that is independent of memory architecture quality. Compounding this concern, competing systems are rapidly approaching the accuracy levels reported here, suggesting that both benchmarks may soon be insufficient to distinguish meaningfully between strong memory architectures. The development of more targeted evaluation protocols, particularly those that stress-test conflict resolution, multi-agent coordination, and non-conversational agentic workflows, represents an important direction for the field.

\textbf{Inference model dependence.} Final results use Gemini~3, contributing $+$4.8\,pp on LongMemEval. As foundational models improve, retrieval quality will likely become an even larger differentiator relative to inference capability.

\textbf{Scale evaluation.} While Moorcheh's engine has been validated at 10M$+$ documents and 2,000$+$ QPS on MAIR \cite{abtahi2025hnswinformationtheoreticbinarizationrethinking}, large scale memory benchmarks testing Memanto with thousands of concurrent agents remain future work.

\textbf{Multi agent memory sharing.} Memanto's namespace architecture currently isolates agent memories by design. Shared memory across agent teams with appropriate access control and consistency protocols is under active development.
\section*{Conclusion}
\label{sec:conc}
We have presented Memanto, a universal memory layer for agentic AI achieving state of the art results on \sys{LongMemEval} (89.8\%) and \sys{LoCoMo} (87.1\%) using a vector only architecture with zero cost ingestion, a 13 category typed semantic memory schema, and built in conflict resolution. A five stage ablation study demonstrated that retrieval recall, rather than architectural complexity, is the dominant performance driver, and that modern LLMs perform the reasoning and filtering that graph based systems attempt to pre compute at ingestion time. By eliminating the Memory Tax, defined as the compounding cost of LLM mediated ingestion, multi query retrieval pipelines, and graph infrastructure management, Memanto enables production grade agentic memory at a fraction of the cost and complexity of hybrid alternatives. Memanto's design embodies a principled trade in which the structural expressiveness of knowledge graphs is exchanged for operational simplicity, determinism, and zero latency ingestion of a single, highly optimised semantic search backend. The empirical results validate this trade decisively.

\bibliographystyle{IEEEtran}
\bibliography{Reference}

@inproceedings{wu2024longmemeval,
  author    = {Di Wu and Hongwei Wang and Wenhao Yu and Yuwei Zhang
               and Kai-Wei Chang and Dong Yu},
  title     = {{LongMemEval}: Benchmarking Chat Assistants on
               Long-Term Interactive Memory},
  booktitle = {Proceedings of the International Conference on
               Learning Representations (ICLR)},
  year      = {2025},
  url       = {https://arxiv.org/abs/2410.10813},
  note      = {arXiv:2410.10813}
}

@inproceedings{maharana2024locomo,
  author    = {Adyasha Maharana and Dong-Ho Lee and Sergey Tulyakov
               and Mohit Bansal and Francesco Barbieri and Yuwei Fang},
  title     = {Evaluating Very Long-Term Conversational Memory of
               {LLM} Agents},
  booktitle = {Proceedings of the 62nd Annual Meeting of the
               Association for Computational Linguistics (ACL)},
  pages     = {13851--13870},
  year      = {2024},
  url       = {https://arxiv.org/abs/2402.17753},
  note      = {arXiv:2402.17753}
}

@misc{chhikara2025mem0buildingproductionreadyai,
      title={Mem0: Building Production-Ready AI Agents with Scalable Long-Term Memory}, 
      author={Prateek Chhikara and Dev Khant and Saket Aryan and Taranjeet Singh and Deshraj Yadav},
      year={2025},
      eprint={2504.19413},
      archivePrefix={arXiv},
      primaryClass={cs.CL},
      url={https://arxiv.org/abs/2504.19413}, 
}

@article{rasmussen2025zep,
  author  = {Preston Rasmussen and Pavlo Paliychuk and Travis Beauvais
             and Jack Ryan and Daniel Chalef},
  title   = {Zep: A Temporal Knowledge Graph Architecture
             for Agent Memory},
  journal = {arXiv preprint arXiv:2501.13956},
  year    = {2025}
}

@misc{packer2024memgptllmsoperatingsystems,
      title={MemGPT: Towards LLMs as Operating Systems}, 
      author={Charles Packer and Sarah Wooders and Kevin Lin and Vivian Fang and Shishir G. Patil and Ion Stoica and Joseph E. Gonzalez},
      year={2024},
      eprint={2310.08560},
      archivePrefix={arXiv},
      primaryClass={cs.AI},
      url={https://arxiv.org/abs/2310.08560}, 
}

@article{xu2025amem,
  author  = {Wujiang Xu and Zujie Liang and Kai Mei and Hang Gao
             and Juntao Tan and Yongfeng Zhang},
  title   = {{A-MEM}: Agentic Memory for {LLM} Agents},
  journal = {arXiv preprint arXiv:2502.12110},
  year    = {2025}
}

@misc{patel2025engram,
      title={ENGRAM: Effective, Lightweight Memory Orchestration for Conversational Agents}, 
      author={Daivik Patel and Shrenik Patel},
      year={2026},
      eprint={2511.12960},
      archivePrefix={arXiv},
      primaryClass={cs.MA},
      url={https://arxiv.org/abs/2511.12960}, 
}

@misc{hindsight2025,
      title={Hindsight is 20/20: Building Agent Memory that Retains, Recalls, and Reflects}, 
      author={Chris Latimer and Nicoló Boschi and Andrew Neeser and Chris Bartholomew and Gaurav Srivastava and Xuan Wang and Naren Ramakrishnan},
      year={2025},
      eprint={2512.12818},
      archivePrefix={arXiv},
      primaryClass={cs.CL},
      url={https://arxiv.org/abs/2512.12818}, 
}

@misc{hu2026memoryageaiagents,
      title={Memory in the Age of AI Agents}, 
      author={Yuyang Hu and Shichun Liu and Yanwei Yue and Guibin Zhang and Boyang Liu and Fangyi Zhu and Jiahang Lin and Honglin Guo and Shihan Dou and Zhiheng Xi and Senjie Jin and Jiejun Tan and Yanbin Yin and Jiongnan Liu and Zeyu Zhang and Zhongxiang Sun and Yutao Zhu and Hao Sun and Boci Peng and Zhenrong Cheng and Xuanbo Fan and Jiaxin Guo and Xinlei Yu and Zhenhong Zhou and Zewen Hu and Jiahao Huo and Junhao Wang and Yuwei Niu and Yu Wang and Zhenfei Yin and Xiaobin Hu and Yue Liao and Qiankun Li and Kun Wang and Wangchunshu Zhou and Yixin Liu and Dawei Cheng and Qi Zhang and Tao Gui and Shirui Pan and Yan Zhang and Philip Torr and Zhicheng Dou and Ji-Rong Wen and Xuanjing Huang and Yu-Gang Jiang and Shuicheng Yan},
      year={2026},
      eprint={2512.13564},
      archivePrefix={arXiv},
      primaryClass={cs.CL},
      url={https://arxiv.org/abs/2512.13564}, 
}

@article{Abou_Ali_2025,
   title={Agentic AI: a comprehensive survey of architectures, applications, and future directions},
   volume={59},
   ISSN={1573-7462},
   url={http://dx.doi.org/10.1007/s10462-025-11422-4},
   DOI={10.1007/s10462-025-11422-4},
   number={1},
   journal={Artificial Intelligence Review},
   publisher={Springer Science and Business Media LLC},
   author={Abou Ali, Mohamad and Dornaika, Fadi and Charafeddine, Jinan},
   year={2025},
   month=nov }

@misc{v2026agenticartificialintelligenceai,
      title={Agentic Artificial Intelligence (AI): Architectures, Taxonomies, and Evaluation of Large Language Model Agents}, 
      author={Arunkumar V and Gangadharan G. R. and Rajkumar Buyya},
      year={2026},
      eprint={2601.12560},
      archivePrefix={arXiv},
      primaryClass={cs.AI},
      url={https://arxiv.org/abs/2601.12560}, 
}

@article{NISA202669,
title = {Agentic AI: The age of reasoning—A review},
journal = {Journal of Automation and Intelligence},
volume = {5},
number = {1},
pages = {69-89},
year = {2026},
issn = {2949-8554},
doi = {https://doi.org/10.1016/j.jai.2025.08.003},
url = {https://www.sciencedirect.com/science/article/pii/S2949855425000516},
author = {Ume Nisa and Muhammad Shirazi and Mohamed Ali Saip and Muhammad Syafiq Mohd Pozi},
}

@inproceedings{yao2023react,
  author    = {Shunyu Yao and Jeffrey Zhao and Dian Yu and Nan Du
               and Izhak Shafran and Karthik Narasimhan
               and Yuan Cao},
  title     = {{ReAct}: Synergizing Reasoning and Acting
               in Language Models},
  booktitle = {Proceedings of ICLR},
  year      = {2023},
  url       = {https://arxiv.org/abs/2210.03629}
}

@article{wang2024survey,
  author  = {Lei Wang and others},
  title   = {A Survey on Large Language Model Based
             Autonomous Agents},
  journal = {Frontiers of Computer Science},
  volume  = {18},
  number  = {6},
  year    = {2024}
}

@article{wei2022chain,
  author  = {Jason Wei and Xuezhi Wang and Dale Schuurmans
             and Maarten Bosma and Brian Ichter
             and Fei Xia and Ed H. Chi and Quoc V. Le
             and Denny Zhou},
  title   = {Chain-of-Thought Prompting Elicits Reasoning
             in Large Language Models},
  journal = {Advances in Neural Information Processing Systems},
  volume  = {35},
  year    = {2022}
}

@article{sumers2023cognitive,
  author  = {Theodore Sumers and Shunyu Yao and Karthik Narasimhan
             and Thomas L. Griffiths},
  title   = {Cognitive Architectures for Language Agents},
  journal = {arXiv preprint arXiv:2309.02427},
  year    = {2023}
}

@article{lewis2020rag,
  author  = {Patrick Lewis and Ethan Perez and Aleksandra Piktus
             and Fabio Petroni and Vladimir Karpukhin
             and Naman Goyal and Heinrich K{\"u}ttler and Mike Lewis
             and Wen-tau Yih and Tim Rockt{\"a}schel and others},
  title   = {Retrieval-Augmented Generation for
             Knowledge-Intensive {NLP} Tasks},
  journal = {Advances in Neural Information Processing Systems},
  volume  = {33},
  pages   = {9459--9474},
  year    = {2020}
}

@article{malkov2018hnsw,
  author  = {Yury A. Malkov and Dmitry A. Yashunin},
  title   = {Efficient and Robust Approximate Nearest Neighbor
             Search Using Hierarchical Navigable Small
             World Graphs},
  journal = {IEEE Transactions on Pattern Analysis and
             Machine Intelligence},
  volume  = {42},
  number  = {4},
  pages   = {824--836},
  year    = {2020},
  doi     = {10.1109/TPAMI.2018.2889473}
}

@article{liu2024lost,
  author  = {Nelson F. Liu and Kevin Lin and John Hewitt
             and Ashwin Paranjape and Michele Bevilacqua
             and Fabio Petroni and Percy Liang},
  title   = {Lost in the Middle: How Language Models Use
             Long Contexts},
  journal = {Transactions of the Association for
             Computational Linguistics},
  volume  = {12},
  pages   = {157--173},
  year    = {2024},
  doi     = {10.1162/tacl_a_00638}
}

@incollection{tulving1972episodic,
  author    = {Endel Tulving},
  title     = {Episodic and Semantic Memory},
  booktitle = {Organization of Memory},
  editor    = {Endel Tulving and Wayne Donaldson},
  pages     = {381--403},
  publisher = {Academic Press},
  year      = {1972}
}

@article{baddeley1992working,
  author  = {Alan D. Baddeley},
  title   = {Working Memory},
  journal = {Science},
  volume  = {255},
  number  = {5044},
  pages   = {556--559},
  year    = {1992},
  doi     = {10.1126/science.1736359}
}

@misc{merrill2026evaluating,
      title={Evaluating Memory Structure in LLM Agents}, 
      author={Alina Shutova and Alexandra Olenina and Ivan Vinogradov and Anton Sinitsin},
      year={2026},
      eprint={2602.11243},
      archivePrefix={arXiv},
      primaryClass={cs.LG},
      url={https://arxiv.org/abs/2602.11243}, 
}

@misc{memoryagentbench2025,
      title={Evaluating Memory in LLM Agents via Incremental Multi-Turn Interactions}, 
      author={Yuanzhe Hu and Yu Wang and Julian McAuley},
      year={2026},
      eprint={2507.05257},
      archivePrefix={arXiv},
      primaryClass={cs.CL},
      url={https://arxiv.org/abs/2507.05257}, 
}

@article{kim2024dialsim,
  author  = {Jiho Kim and Woosog Chay and Hyeonji Hwang
             and Daeun Kyung and Hyunseung Chung
             and Eunbyeol Cho and Yohan Jo and Edward Choi},
  title   = {{DialSim}: A Real-Time Simulator for Evaluating
             Long-Term Dialogue Understanding of Conversational
             Agents},
  journal = {arXiv preprint arXiv:2406.13144},
  year    = {2024}
}

@misc{macpherson2025episodic,
      title={Position: Episodic Memory is the Missing Piece for Long-Term LLM Agents}, 
      author={Mathis Pink and Qinyuan Wu and Vy Ai Vo and Javier Turek and Jianing Mu and Alexander Huth and Mariya Toneva},
      year={2025},
      eprint={2502.06975},
      archivePrefix={arXiv},
      primaryClass={cs.AI},
      url={https://arxiv.org/abs/2502.06975}, 
}

@misc{hipporag2024,
      title={HippoRAG: Neurobiologically Inspired Long-Term Memory for Large Language Models}, 
      author={Bernal Jiménez Gutiérrez and Yiheng Shu and Yu Gu and Michihiro Yasunaga and Yu Su},
      year={2025},
      eprint={2405.14831},
      archivePrefix={arXiv},
      primaryClass={cs.CL},
      url={https://arxiv.org/abs/2405.14831}, 
}

@inproceedings{le2025reflective,
    title = "In Prospect and Retrospect: Reflective Memory Management for Long-term Personalized Dialogue Agents",
    author = "Tan, Zhen  and
      Yan, Jun  and
      Hsu, I-Hung  and
      Han, Rujun  and
      Wang, Zifeng  and
      Le, Long  and
      Song, Yiwen  and
      Chen, Yanfei  and
      Palangi, Hamid  and
      Lee, George  and
      Iyer, Anand Rajan  and
      Chen, Tianlong  and
      Liu, Huan  and
      Lee, Chen-Yu  and
      Pfister, Tomas",
    editor = "Che, Wanxiang  and
      Nabende, Joyce  and
      Shutova, Ekaterina  and
      Pilehvar, Mohammad Taher",
    booktitle = "Proceedings of the 63rd Annual Meeting of the Association for Computational Linguistics (Volume 1: Long Papers)",
    month = jul,
    year = "2025",
    address = "Vienna, Austria",
    publisher = "Association for Computational Linguistics",
    url = "https://aclanthology.org/2025.acl-long.413/",
    doi = "10.18653/v1/2025.acl-long.413",
    pages = "8416--8439",
    ISBN = "979-8-89176-251-0"
}

@misc{raptor2024,
      title={RAPTOR: Recursive Abstractive Processing for Tree-Organized Retrieval}, 
      author={Parth Sarthi and Salman Abdullah and Aditi Tuli and Shubh Khanna and Anna Goldie and Christopher D. Manning},
      year={2024},
      eprint={2401.18059},
      archivePrefix={arXiv},
      primaryClass={cs.CL},
      url={https://arxiv.org/abs/2401.18059}, 
}

@inproceedings{shi2024replug,
  author    = {Weijia Shi and Sewon Min and Michihiro Yasunaga
               and Minjoon Seo and Richard James and Mike Lewis
               and Luke Zettlemoyer and Wen-tau Yih},
  title     = {{REPLUG}: Retrieval-Augmented Black-Box
               Language Models},
  booktitle = {Proceedings of NAACL},
  pages     = {8371--8384},
  year      = {2024}
}

@inproceedings{zhong2024memorybank,
  title={Memorybank: Enhancing large language models with long-term memory},
  author={Zhong, Wanjun and Guo, Lianghong and Gao, Qiqi and Ye, He and Wang, Yanlin},
  booktitle={Proceedings of the AAAI conference on artificial intelligence},
  volume={38},
  number={17},
  pages={19724--19731},
  year={2024}
}

@inproceedings{du2024perltqa,
  author    = {Yiming Du and Hongru Wang and Zhengyi Zhao
               and Bin Liang and Baojun Wang and Wanjun Zhong
               and Zezhong Wang and Kam-Fai Wong},
  title     = {{PerLTQA}: A Personal Long-Term Memory Dataset
               for Memory Classification, Retrieval, and Fusion
               in Question Answering},
  booktitle = {Proceedings of SIGHAN 2024},
  pages     = {152--164},
  year      = {2024}
}

@misc{abtahi2025hnswinformationtheoreticbinarizationrethinking,
      title={From HNSW to Information-Theoretic Binarization: Rethinking the Architecture of Scalable Vector Search}, 
      author={Seyed Moein Abtahi and Majid Fekri and Tara Khani and Akramul Azim},
      year={2025},
      eprint={2601.11557},
      archivePrefix={arXiv},
      primaryClass={cs.DB},
      url={https://arxiv.org/abs/2601.11557}, 
}

@misc{terranova2025evaluatinglongtermmemorylongcontext,
      title={Evaluating Long-Term Memory for Long-Context Question Answering}, 
      author={Alessandra Terranova and Björn Ross and Alexandra Birch},
      year={2025},
      eprint={2510.23730},
      archivePrefix={arXiv},
      primaryClass={cs.CL},
      url={https://arxiv.org/abs/2510.23730}, 
}

@techreport{claude2025,
  author       = {Anthropic},
  title        = {Claude Model Card},
  institution  = {Anthropic},
  year         = {2026},
  howpublished = {\url{https://www.anthropic.com/claude-3-model-card}},
  note         = {Accessed: 2026}
}
\appendix
All benchmark evaluations are fully reproducible using the configuration described below. Code and evaluation scripts are publicly available on GitHub\footnote{\url{https://github.com/moorcheh-ai/memanto-evaluation}}, and datasets are hosted on Hugging Face\footnote{\url{https://huggingface.co/moorcheh}}.

\subsection{Software Configuration}

All benchmark evaluations are conducted using Memanto version 2.1.4 in conjunction with the latest production release of the Moorcheh SDK. Claude Sonnet~4 serves as the inference model for Stages~1 through~4, while Gemini~3 is employed at Stage~5 to establish parity with leading competing systems. Evaluation prompts for both answer generation and LLM judging are inspired by and partially adapted from the Hindsight repository~\cite{hindsight2025}; however, because Memanto's retrieval context is structurally different from Hindsight's chunk-plus-reflection pipeline, the prompts have been modified accordingly. The \sys{LongMemEval} assessment uses the full 500-question suite under the standard $S$ setting, and \sys{LoCoMo} is evaluated on its standard split. Claude Sonnet~4 serves as the LLM judge throughout all evaluation stages.

\subsection{Hardware Requirements}

All experiments require a minimum of four CPU cores and 8\,GB of RAM, with 16\,GB recommended for parallel evaluation workloads. Approximately 10\,GB of disk space is required to accommodate the benchmark datasets. GPU access is not a prerequisite, as inference is performed via the managed Gemini~3 API. The software environment requires Python~3.10 or later, with Docker employed for containerised deployment.

\subsection{Memory Type Assignment}

Memory type assignment is currently performed by the user at write time, who selects the appropriate category from the 13-type schema described in Table~\ref{tab:schema}. Automated type assignment via a rule-based decision tree is planned for a future release and will eliminate this manual step entirely.

\subsection{Final Retrieval Parameters}

The final evaluation configuration employs a dynamic retrieval budget of up to 100 chunks, governed by ITS threshold gating rather than a fixed top-$k$ constraint. The ITS similarity threshold is set to 0.05, and a single retrieval query is issued per question, with no multi-query or recursive retrieval strategies applied.
\end{document}